%% file: main.tex
\documentclass[10pt,twocolumn,letterpaper]{article}

\usepackage[pagenumbers]{cvpr} % To force page numbers, e.g. for an arXiv version

\usepackage{hyperref}

\usepackage{multirow}
\usepackage{amsmath}
\usepackage{amsfonts}
\usepackage{algorithm}
\usepackage{algorithmic}

\def\paperID{*****} % *** Enter the Paper ID here
\def\confName{CVPR}
\def\confYear{2024}

\title{FiLo: Zero-Shot Anomaly Detection by Fine-Grained Description and High-Quality Localization}

\author{
Zhaopeng Gu$^{1,2}$~~~~
Bingke Zhu$^{1,3}$~~~~
Guibo Zhu$^{1,2}$~~~~
Yingying Chen$^{1,3}$~~~~\\
Hao Li$^{4}$\thanks{Work done as intern in CASIA.}~~~~
Ming Tang$^{1,2}$~~~~
Jinqiao Wang$^{1,2,3}$\\
  $^{1}$~Foundation Model Research Center, Institute of Automation, \\ Chinese Academy of Sciences, Beijing, China \\ 
  $^{2}$~University of Chinese Academy of Sciences, Beijing, China\\
  $^{3}$~Objecteye Inc., Beijing, China\\
  $^{4}$~Central South University, Hunan, China \\
  {\tt\small  guzhaopeng2023@ia.ac.cn} \\
  {\tt\small \{bingke.zhu,gbzhu,yingying.chen,tangm,jqwang\}@nlpr.ia.ac.cn} \\
  {\tt\small 8209210109@csu.edu.cn}
}

\begin{document}

\definecolor{cvprblue}{rgb}{0.21,0.49,0.74}
\hypersetup{pagebackref,breaklinks,colorlinks,citecolor=cvprblue}
% \vspace{-8mm}
\maketitle
\input{sec/0_abstract}    
\input{sec/1_intro}
\input{sec/2_related}
\input{sec/3_method}

\input{sec/4_experiment}

\input{sec/5_conclusion}
{
    \small
    \bibliographystyle{ieeenat_fullname}
    \bibliography{main}
}

\input{sec/6_suppl}

% {
%     \small
%     \bibliographystyle{ieeenat_fullname}
%     \bibliography{main}
% }


\end{document}

%% file: sec/0_abstract.tex
\begin{abstract}
  Zero-shot anomaly detection (ZSAD) methods detect anomalies without prior access to known normal or abnormal samples within target categories. Existing methods typically rely on pretrained multimodal models, computing similarities between manually crafted textual features representing "normal" or "abnormal" semantics and image patch features to detect anomalies. However, the generic descriptions of "abnormal" often fail to precisely match diverse types of anomalies across different object categories. Additionally, computing feature similarities for single patches struggles to pinpoint specific locations of anomalies with various sizes and scales. To address these issues, we propose a novel ZSAD method called FiLo, comprising two components: adaptively learned \textbf{F}ine-\textbf{G}rained \textbf{Des}cription~(FG-Des) and position-enhanced \textbf{H}igh-\textbf{Q}uality \textbf{Loc}alization~(HQ-Loc). FG-Des introduces fine-grained anomaly descriptions for each category using Large Language Models~(LLMs) and employs adaptively learned textual templates to enhance the accuracy and interpretability of anomaly detection. HQ-Loc, utilizing Grounding DINO for preliminary localization, position-enhanced text prompts, and Multi-scale Multi-shape Cross-modal Interaction~(MMCI) module, facilitates more accurate localization of anomalies of different sizes and shapes. Experimental results on datasets like MVTec and VisA demonstrate that FiLo significantly improves the performance of ZSAD in both detection and localization, achieving state-of-the-art performance with an image-level AUC of 83.9\% and a pixel-level AUC of 95.9\% on the VisA dataset.  Code is available at https://github.com/CASIA-IVA-Lab/FiLo.
\end{abstract}

%% file: sec/1_intro.tex
\section{Introduction}
% \vspace{-2mm}
\begin{figure}[t]
  \centering
  \includegraphics[width=0.95\linewidth]{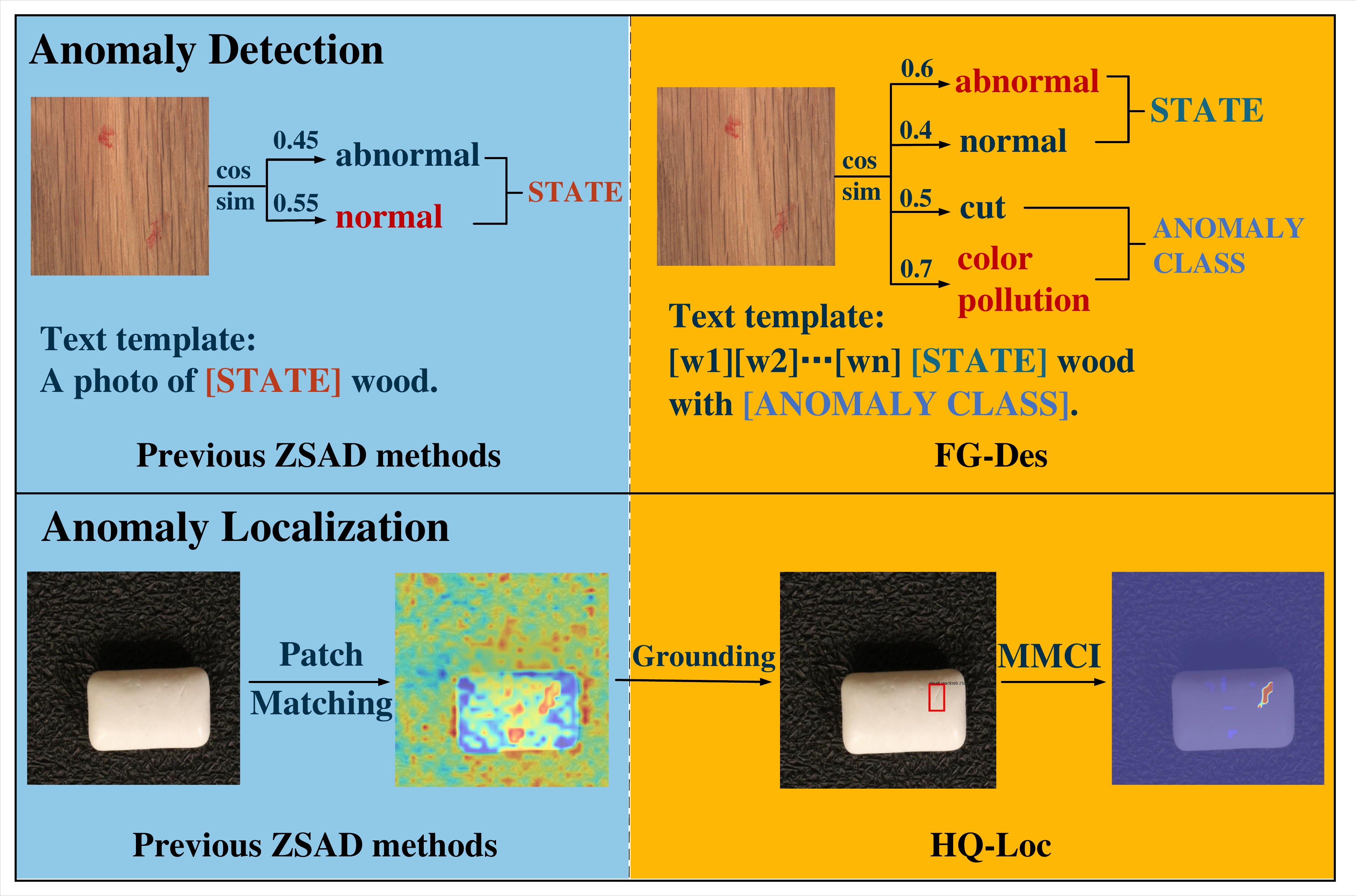} 
  \caption{Comparison of anomaly detection and localization between FiLo and previous ZSAD methods. Previous ZSAD methods utilize fixed templates and generic anomaly descriptions, potentially resulting in errors. Our FG-Des enhances detection accuracy with adaptively learned text templates and fine-grained anomaly descriptions. For localization, ZSAD methods often produce false positives in background areas by directly comparing image patches with text features. Our HQ-Loc approach, using Grounding DINO, location enhancement, and MMCI, effectively removes background regions and improves localization accuracy.}
  \label{fig:compare}
\end{figure}

The anomaly detection task aims to identify whether industrial products contain abnormalities or defects and locate the abnormal regions within the samples, which plays a crucial role in product quality control and safety monitoring. Traditional anomaly detection methods~\cite{roth2022towards, you2022unified, deng2022anomaly, defard2021padim} typically require a large number of normal samples for model training. While performing well in some scenarios, they often fail in situations requiring protection of user data privacy or when applied to new production lines. Zero-Shot Anomaly Detection~(ZSAD) has emerged as a research direction tailored to such scenarios, aiming to perform anomaly detection tasks without prior data on the target item categories, demanding high generalization ability from the model.

Multimodal pre-trained models~\cite{radford2021learning, kirillov2023segment, li2023blip} have recently demonstrated strong zero-shot recognition capabilities in various visual tasks. Many works have sought to leverage the vision-language comprehension ability of multimodal pre-trained models for ZSAD tasks, such as WinCLIP~\cite{jeong2023winclip}, APRIL-GAN~\cite{chen2023zero}, and AnomalyGPT~\cite{gu2024anomalygpt}. These methods assess whether an image contains anomalies by computing the similarity between image features and manually crafted textual features representing "normal" and "abnormal" semantics. They also localize abnormal regions by calculating the similarity between the image patch features and the textual features. While these approaches partly address the challenges of ZSAD, they encounter issues in both anomaly detection and localization. The generic "abnormal" descriptions fail to precisely match the diverse types of anomalies across different object categories. Moreover, computing feature similarity for individual patches struggles to precisely locate abnormal regions of varying sizes and shapes. To tackle these issues, we propose FiLo~(\textbf{Fi}ne-Grained Description and High-Quality \textbf{Lo}calization), which addresses the shortcomings of existing ZSAD methods through adaptively learned Fine-Grained Description (FG-Des) and High-Quality Localization (HQ-Loc), as depicted in Figure~\ref{fig:compare}.

Concerning anomaly detection, manually crafted abnormal descriptions typically employ generic terms such as "damaged" or "defect"~\cite{jeong2023winclip, gu2024anomalygpt, chen2023zero}, which do not adequately capture the specific types of anomalies present across different object categories. Furthermore, existing methods' text prompt templates like \textit{A xxx photo of xxx.} are primarily designed for foreground object classification tasks and may not be suitable for identifying normal and abnormal parts within objects. In FG-Des, we first leverage the capabilities of Large Language Models (LLMs) to generate fine-grained anomaly types for each object category, replacing generic abnormal descriptions with specific anomaly content that matches the anomaly samples better. Next, we utilize learnable text vectors instead of manually crafted sentence templates and embed the detailed anomaly content generated in the previous step into the adaptively learned text templates to improve the match between the text and the abnormal images, enhancing the textual features for anomaly detection. Our FG-Des not only improves the accuracy of anomaly detection but also enables the identification of the specific anomaly categories present in the samples, thus enhancing the interpretability.

Regarding anomaly localization, existing methods~\cite{gu2024anomalygpt, chen2023zero, deng2023anovl} localize anomalies by computing the similarity between the features of each image patch and the textual features. However, anomalies often span multiple patches with different shapes and sizes, sometimes requiring comparison with surrounding normal regions to determine their abnormality. While WinCLIP~\cite{jeong2023winclip} addresses this issue by employing windows of different sizes, it incurs significant time and space costs by inputting a large number of images corresponding to each window into CLIP's image encoder during inference. To tackle this problem, we design HQ-Loc, which consists of three main components: first, preliminary anomaly localization based on Grounding DINO~\cite{liu2023grounding}. Considering that even in abnormal samples, most regions are normal, and anomalies only exist in small local areas, we utilize the detailed anomaly descriptions generated in the previous step and employ Grounding DINO~\cite{liu2023grounding} for preliminary anomaly localization. Although directly using Grounding DINO for zero-shot anomaly localization yields low accuracy, the localized regions are always in the foreground, effectively avoiding false positives in background regions. Second, position enhancement involves adding the position detected by Grounding DINO to the text prompt, resulting in a more accurate description of the anomaly position. Third, the Multi-scale Multi-shape Cross-modal Interaction (MMCI) module aggregates patch features extracted by the Image Encoder using convolutional kernels of different sizes and shapes to enhance the method's ability to localize anomalies of different sizes and shapes.

Extensive experiments are conducted on multiple datasets like MVTec~\cite{bergmann2019mvtec} and VisA~\cite{zou2022spot}. Our FiLo improves the accuracy of anomaly detection and localization, achieving new state-of-the-art zero-shot performance. Trained on the MVTec dataset and tested on the VisA dataset, FiLo achieves an image-level AUC of 83.9\% and a pixel-level AUC of 95.9\%, outperforming other ZSAD methods.

Our contributions can be summarized as follows:
\begin{itemize}
\item We propose an adaptively learned Fine-Grained Description (FG-Des) that leverages domain-specific knowledge to introduce detailed anomaly descriptions, replacing generic "normal" and "abnormal" descriptions. Also, we use learnable vectors instead of manually crafted text templates to learn textual content which is more suitable for anomaly detection task, improving both the accuracy and interpretability.
\item Additionally, we design a position-enhanced High-Quality Localization method (HQ-Loc) that employs Grounding DINO~\cite{liu2023grounding} for preliminary anomaly localization, enhances text prompts with descriptions of anomaly positions, and utilizes an MMCI module to localize anomalies of different sizes and shapes more accurately, improving anomaly localization accuracy.
\item Experiments on multiple datasets demonstrate significant performance improvements in anomaly detection and localization compared to baseline methods. FiLo has been proved to be effective for zero-shot anomaly detection and localization, achieving state-of-the-art performance.
\end{itemize}

%% file: sec/2_related.tex
\section{Related work}
\subsection{Vision-Language Models}
Recently, multimodal models integrating visual and textual content have achieved significant success in various visual tasks~\cite{radford2021learning, li2023blip, liu2023grounding}. Among these, CLIP~\cite{radford2021learning}, pre-trained on a massive scale internet dataset, emerges as one of the most prominent methods. CLIP employs two structurally similar Transformer~\cite{vaswani2017attention} encoders to extract features from images and text, aligning features with the same semantics through contrastive learning methods. With appropriate prompts, CLIP demonstrates remarkable zero-shot generalization capabilities across multiple datasets for downstream image classification tasks. However, the quality of prompts significantly affects the performance of downstream tasks. Traditional approaches~\cite{jeong2023winclip, cao2023segment} require experts to manually craft suitable text prompts for each task, demanding domain-specific knowledge and being time-consuming. Recent methods like coop~\cite{zhou2022learning} and cocoop~\cite{zhou2022conditional} propose using learnable vectors instead of manually crafted prompts, requiring minimal training cost while achieving superior performance across multiple datasets.

While the original CLIP was designed for image classification tasks, researchers have extended their efforts to explore vision-language models for object detection and semantic segmentation tasks. Grounding DINO~\cite{liu2023grounding} is a notable example, combining the Transformer-based object detector DINO with Grounded pretraining, achieving excellent performance as an open-set object detector.

Our FG-Des method, incorporating adaptive learned fine-grained anomaly descriptions, is built upon CLIP~\cite{radford2021learning} and cocoop~\cite{zhou2022conditional}. However, straightforward utilization of cocoop-enhanced CLIP does not excel in anomaly detection tasks. Detailed anomaly descriptions for each item category are crucial for achieving outstanding performance. Grounding DINO~\cite{liu2023grounding} serves as a vital component of HQ-Loc. Yet, employing Grounding DINO~\cite{liu2023grounding} directly for zero-shot anomaly localization yields low accuracy. We utilize Grounding DINO solely for preliminary anomaly localization, capturing the approximate location of anomalies and avoiding false positives in background regions.

\subsection{Zero-shot Anomaly Detection}
Most zero-shot anomaly detection methods leverage the transferability of pre-trained vision-language models. Early methods like ZoC~\cite{esmaeilpour2022zero} and CLIP-AD~\cite{liznerski2022exposing}, simply apply CLIP to anomaly detection data, resulting in low accuracy and inability to localize abnormal regions. WinCLIP~\cite{jeong2023winclip} first achieves anomaly localization by cropping windows of different sizes in images and significantly enhances anomaly detection by employing carefully crafted text prompts. APRIL-GAN~\cite{chen2023zero} aligns patch-level image features with textual features using a learnable linear projection layer to accomplish anomaly localization, overcoming the inefficiency caused by WinCLIP's input of numerous windows and further enhancing performance. AnoVL~\cite{deng2023anovl} resolves the mismatch between patch-level image features and textual features by introducing V-V attention~\cite{li2023clip}, enabling direct application of CLIP to anomaly detection tasks without any additional training. However, all the above methods require carefully designed and manually crafted text templates. AnomalyCLIP~\cite{zhou2023anomalyclip}, an emerging approach, substitutes object-agnostic learnable text vectors for manually crafted text templates. Nevertheless, AnomalyCLIP describes anomalies uniformly using the word "damaged", which is evidently insufficient to cover all types of anomalies. 

Segment Any Anomay~(SAA)~\cite{cao2023segment} is a zero-shot anomaly localization method based on the Grounded-SAM~\cite{ren2024grounded} approach. SAA utilizes Grounding DINO to generate anomaly bounding boxes, which are then used as prompts input into the Segment Anything Model~\cite{kirillov2023segment} to obtain anomaly localization results. However, SAA~\cite{cao2023segment} requires expertly crafted text inputs for Grounding DINO, and its results heavily rely on the detection outcomes of Grounding DINO, which may lead to low precision when directly applied to ZSAD. In our method, Grounding DINO serves solely as a preliminary anomaly localization module, aiming to prevent false positives in background regions of images. The primary dependency of our approach lies in the MMCI module for anomaly localization.

Moreover, none of the above methods incorporate location information of anomalies in the text prompt. Compared to existing methods, our approach enhances anomaly detection performance and interpretability by adaptive learned Fine-Grained anomaly Descriptions. We also improve the localization capability for anomalies of different sizes and shapes through our position-enhanced High-Quality localization method HQ-Loc.

\subsection{Visual Description Enhancement}
Numerous prior studies~\cite{zhou2022conditional,zhou2022learning} have extensively demonstrated that the quality of the text prompt significantly impacts the performance of downstream tasks for pretrained Vision-Language models like CLIP~\cite{radford2021learning}. In contrast to text content meticulously crafted by experts, recent works~\cite{maniparambil2023enhancing, menon2022visual,feng2023leveraging} have delegated the task of generating high-quality text prompts to LLMs, which are called visual description enhancement. LLMs such as GPT-3.5~\cite{ouyang2022training} and GPT-4~\cite{achiam2023gpt} encapsulate extensive knowledge across various domains, showcasing impressive performance across a spectrum of tasks. FiLo harnesses the profound domain knowledge embedded within LLMs to generate potential anomaly types for each item category, deriving fine-grained anomaly descriptions. We are the first to apply visual description enhancement techniques to anomaly detection tasks.

\subsection{Multi-Scale Convolution}
In recent years, multi-scale convolution has been a research hotspot to detect objects of different sizes appearing in images~\cite{szegedy2015going, ding2021repvgg, tan2019mixconv, ding2019acnet}. Multi-scale convolution methods aggregate features of regions with different sizes by using convolutional kernels of various sizes, achieving significant performance improvements in image classification, semantic segmentation, and object detection. InceptionNet~\cite{szegedy2015going} is a typical representative, simultaneously employing convolutional kernels of $1\times 1$, $3\times 3$, $5\times 5$, etc. within the same layer to address the uncertainty of the optimal kernel size across different samples. MixConv~\cite{tan2019mixconv} groups input channels and applies convolutional kernels of different sizes to each channel group. RepVGG~\cite{ding2021repvgg} decomposes all sizes of convolutional kernels into a series of composite operations of $3\times 3$ convolutions. ACNet~\cite{ding2019acnet} changes the order of convolution and summation, first summing convolutional kernels of different sizes and then performing a single convolution operation, thereby reducing computational overhead. Most existing multi-scale methods focus on square convolutional kernels of different sizes. ACNet~\cite{ding2019acnet} employs multi-shape convolutional kernels, but its emphasis is on computational efficiency, neglecting multi-scale aspects. Since anomalies in images may exhibit various shapes and sizes, our MMCI module introduces convolutional kernels of different sizes and shapes to fully localize anomalies.

%% file: sec/3_method.tex
\begin{figure*}[t]
  \centering
  \includegraphics[width=0.98\textwidth]{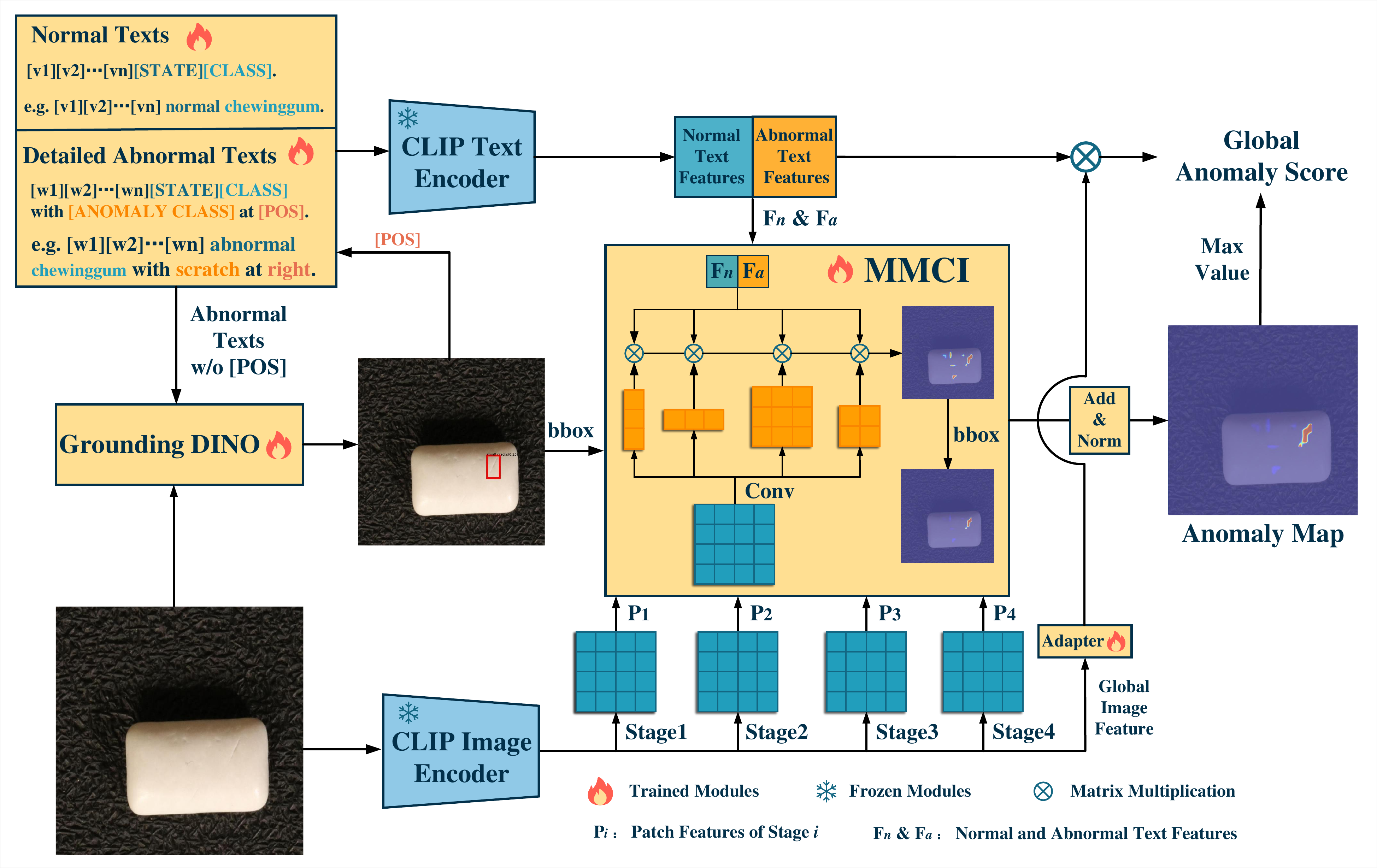} 
  \caption{Overall architecture of FiLo. Given an input image, fine-grained anomaly types are generated by LLM. Then normal and detailed abnormal texts are input into Grounding DINO to obtain bounding boxes and are fed into CLIP Text Encoder to get $F_n$ and $F_a$. Intermediate patch features of input image are subjected to MMCI together with text features to compute anomaly map, and the global image features are compared with text features after adaptation to obtain global anomaly score.}
  \label{fig:arch}
\end{figure*}

\section{FiLo}

In this paper, we propose a vision-language ZSAD method, FiLo, to enhance the capability of zero-shot anomaly detection and localization. Regarding anomaly detection, we devise the adaptively learned Fine-Grained Description method (FG-Des, Sec 3.2), which leverages fine-grained anomaly descriptions generated by LLMs and adaptable text vectors to identify the most precise textual representation for each anomaly sample. FG-Des facilitates more accurate judgments regarding the presence of anomalies in images and determines detailed anomaly types, thereby enhancing the interpretability of the method. For anomaly localization, we introduce the position-enhanced High-Quality Localization method (HQ-Loc, Sec 3.3), which employs preliminary localization via Grounding DINO, position-enhanced text prompts, and a Multi-scale, Multi-shape Cross-modal Interaction module to more accurately pinpoint anomalies of various sizes and shapes.

\subsection{Overall Architecture}

The overall architecture of the model is illustrated in Figure~\ref{fig:arch}. For an input image $I\in \mathbb{R}^{H\times W\times 3}$, we first utilize information from the dataset or LLM to generate a list of fine-grained anomaly types that may exist for this item category. Subsequently, the anomaly text is inputted into Grounding DINO to obtain preliminary bounding boxes for anomaly localization. Simultaneously, the combination of fine-grained anomaly type and previously learned text vector templates yields text descriptions for both normal and abnormal cases. These descriptions are then fed into the CLIP Text Encoder for feature extraction, resulting in representations of normal and abnormal text features. Next, the image is passed through the CLIP Image Encoder to extract intermediate patch features $P_i \in \mathbb{R}^{H_i\times W_i\times C_i}$ from M stages, where $i$ indicates the $i$-th stage. These intermediate patch features are subjected to the MMCI module together with text features to generate anomaly map for each layer $M_i \in \mathbb{R}^{H\times W}$. Subsequently, after filtering with bounding boxes, the score maps for each layer are summed and normalized to obtain the final anomaly map $M \in \mathbb{R}^{H\times W}$. The global features of the image are compared with text features after adaptation, and the maximum value of the final anomaly map $M$ is added to derive the global anomaly score for the image.

\subsection{FG-Des}
Numerous existing methods~\cite{jeong2023winclip,deng2023anovl,chen2023zero} have demonstrated that the quality of text prompts significantly affects the effectiveness of anomaly detection when performing zero-shot inference on new categories. Therefore, we first focus on prompt engineering to generate more accurate and efficient text prompts for enhancing anomaly detection in ZSAD. In FG-Des, we achieve this goal through adaptively learned text templates and fine-grained anomaly descriptions generated by LLMs. 
% The following is a detailed explanation of these two components.

\subsubsection{Adaptively Learned Text Templates}

Following the success of methods like WinCLIP~\cite{jeong2023winclip}, subsequent methods such as APRIL-GAN~\cite{chen2023zero} and AnomalyGPT~\cite{gu2024anomalygpt} directly adopt the text templates used in WinCLIP to construct text prompts. However, the text template in WinCLIP, \textit{A xxx photo of [state] [class]}, is primarily derived from the text template used by CLIP for image classification tasks on the ImageNet~\cite{deng2009imagenet} dataset, which mainly indicates the category of foreground objects in the image rather than whether the object contains anomalies internally. To address this issue, we employ adaptive text templates learned based on anomaly detection-related data. During the learning process, these templates can combine the normal and abnormal content in the image to generate text prompts that better distinguish between normal and abnormal cases, while avoiding the need for extensive manual template engineering. Our adaptive normal and abnormal text templates are defined as follows:

\begin{equation*}
\begin{aligned}
    T_n\ =\ &[V_1][V_2]...[V_n][STATE][CLASS]. \\
    T_a\ =\ &[W_1][W_2]...[W_n][STATE][CLASS] \\
          & with\ [ANOMALY\ CLASS]\ at\ [POS].
\end{aligned}
\end{equation*}

\noindent $[V_i]$ and $[W_i]$ are learnable text vectors, $[STATE]$ represents the general "normal" or "abnormal" state, $[CLASS]$ denotes the item category, $[ANOMALY\ CLASS]$ specifies the detailed anomaly content, and $[POS]$ indicates the location of the anomaly region, which can be one of nine possible scenarios, e.g., "top left" or "bottom".

Based on this template, we only need to replace the $[CLASS]$, $[ANOMALY\ CLASS]$, and $[POS]$ parts for different objects to generate different text prompt content.

\subsubsection{Fine-Grained Anomaly Descriptions}

As mentioned earlier, the generic "anomaly" texts in existing methods are insufficient to accurately describe the diverse types of anomalies that may appear on different object categories. Therefore, there is an urgent need for more personalized, informative text prompts to accurately characterize each image. LLMs such as GPT-4~\cite{achiam2023gpt} possess rich expert knowledge across various domains. We harness the power of LLMs to generate specific lists of potential anomaly types for each item category, replacing the vague and general "anomaly" or "damaged" descriptions used in previous methods. Such detailed textual features, when combined with features extracted by CLIP from images, lead to better anomaly detection results.

By incorporating fine-grained anomaly descriptions generated by large language models (LLMs) into the adaptive text templates' $[ANOMALY\ CLASS]$ section, we obtain complete text prompts. These prompts are then inputted into the CLIP Text Encoder, and after group averaging, we obtain text features representing normal and abnormal cases, denoted as $F = [F_n,F_a] \in \mathbb{R}^{2\times C}$. For the global features $G$ extracted from the image via the CLIP Image Encoder, we first pass them through a linear adapter layer to obtain adapted image features $A\in \mathbb{R}^{C}$ that better match the textual content. Next, we calculate the global anomaly score by Eq~(\ref{eq:s_global}):

\begin{equation}
    S_{global} = softmax(A\cdot F_a^T) + \max(M).
    \label{eq:s_global}
\end{equation}
$M$ represents the anomaly map calculated in Sec~3.3 and max$(\cdot)$ denotes the maximum operation.

Fine-grained anomaly descriptions not only improve the accuracy of anomaly detection but also enhance the interpretability of the detection results. Specifically, we can calculate the similarity between image features and each precise anomaly description. By examining the textual descriptions with high similarity, we can determine which category the anomaly in the image belongs to, thus gaining deeper insight into the model's decision-making process.

\subsection{HQ-Loc}
Existing Zero-Shot Anomaly Detection (ZSAD) methods often locate anomaly positions by computing the similarity between the features of each image patch and textual features. However, an anomaly region often spans multiple patches, exhibiting various positions, shapes, and sizes. Sometimes, it requires comparison with surrounding normal regions to determine if it's an anomaly. To address this, we propose this position-enhanced High-Quality Localization method HQ-Loc, which enhances anomaly localization from coarse to fine. This is achieved through three key components: Grounding DINO preliminary localization, position-enhanced textual prompts, and Multi-Scale Multi-Shape Cross-modal Interaction Module (MMCI). Below, we provide detailed explanations for each component.

\subsubsection{Grounding DINO Preliminary Localization}
Existing ZSAD methods typically lack discrimination between patches at different positions in the image, often resulting in the misidentification of background perturbations as anomalies. To mitigate this, we utilize detailed anomaly descriptions generated in the previous step to perform preliminary anomaly localization using Grounding DINO. While direct application of Grounding DINO may not precisely determine the exact location of anomalies, the localization boxes obtained generally reside in the foreground of objects, often near the anomaly area. Therefore, using the localization results from Grounding DINO to restrict anomaly regions effectively avoids false positives in the background, thus enhancing the accuracy of anomaly localization. Additionally, since Grounding DINO localization is not entirely accurate and may have missed detections, we adopt a strategy of suppressing anomaly scores outside all boxes by multiplying them with a parameter $\lambda$.

\subsubsection{Position-Enhanced Textual Prompt}
After obtaining the preliminary anomaly localization results from Grounding DINO, we incorporate the position information from the localization boxes into textual prompts to enhance position descriptions. Textual prompts with detailed anomaly descriptions and position enhancements are more aligned with the content in the image being examined. This alignment assists the model in concentrating on specific areas of the image during anomaly localization in the subsequent step, thereby improving localization accuracy.

\subsubsection{MMCI Module}
To comprehensively locate anomalies of different shapes and sizes, our approach does not directly compute the similarity between each image patch feature and textual features. Instead, we design a Multi-Scale Multi-Shape Cross-Modal Interaction Module (MMCI). MMCI is inspired by WinCLIP's use of windows of different sizes to select subregions in images and then determine if each subregion contains an anomaly. However, MMCI significantly reduces the computational overhead incurred by WinCLIP when simultaneously inputting dozens of images selected by windows into the CLIP's Image Encoder. Specifically, we design convolutional kernels of different sizes and shapes to process patch features extracted by the CLIP Image Encoder in parallel. Subsequently, we aggregate these features and compute their similarity with position-enhanced textual features. Through this approach, our MMCI module can effectively handle anomalies of different sizes and shapes, enhancing the model's ability to localize anomaly regions.

Let $n$ different shaped convolutional kernels be denoted as $C_j$, where $j$ ranges from $1$ to $n$. Given patch features $P_i \in \mathbb{R}^{H_i W_i\times C}$, position-enhanced text features $[F_n, F_a]\in \mathbb{R}^{2\times C}$, normal map $M^n_i \in \mathbb{R}^{H\times W}$ 
 and anomaly map $M^a_i \in \mathbb{R}^{H\times W}$ can be calculated by Eq.~(\ref{eq:m_i}):

\begin{equation}
    M^n_i, M^a_i = Up(Norm(\sum_{j=1}^{n} S(C_j(P_i)\cdot [F_n, F_a]^T))),
    \label{eq:m_i}
\end{equation}

\noindent where $Up(\cdot)$ denotes the upsampling operation, $S(\cdot)$ is the softmax operation, and $Norm(\cdot)$ represents the normalization operation, ensuring that the values in the anomaly map lie between 0 and 1.
By summing and normalizing $M_i$ for each layer, we can obtain the normal and anomaly map:
\begin{equation}
    M^n = Norm(\sum_i {M^n_i}),\ M^a = Norm(\sum_i {M^a_i}),
\end{equation}
and the final localization result can be calculated by Eq~(\ref{eq:final_m})
\begin{equation}
    M = G_\sigma (M^a + 1 - M^n) / 2, \label{eq:final_m}
\end{equation}
where $G_\sigma$ is a Gaussian filter, and $\sigma$ controls smoothing.

\subsection{Adapter}
We employ a common bottleneck structure Adapter to align global image features and text features, consisting of two linear layers, one ReLU~\cite{glorot2011deep} layer, and one SiLU~\cite{elfwing2018sigmoid} layer, as shown in Algorithm~\ref{alg:adapter}.

% \begin{figure}[h]
%   \centering
%   \includegraphics[width=0.5\linewidth]{figs/adapter.pdf} 
%   \caption{Architecture of adapter module.}
%   \label{fig:adapter}
% \end{figure}

\begin{algorithm}[h]
\caption{Adapter Module}
\label{alg:adapter}
\begin{algorithmic}[1]
\REQUIRE Input vector $\mathbf{x} \in \mathbb{R}^{768}$
\ENSURE Output vector $\mathbf{y} \in \mathbb{R}^{768}$
\STATE $\mathbf{h}_1 = \text{ReLU}(\mathbf{W}_1 \mathbf{x} + \mathbf{b}_1) \in \mathbb{R}^{384}$ 
\STATE $\mathbf{y} = \text{SiLU}(\mathbf{W}_2 \mathbf{h}_1 + \mathbf{b}_2)$ 
\end{algorithmic}
\end{algorithm}

\subsection{Loss Functions}
To learn the content of adaptive text templates and the convolutional kernel parameters in MMCI, we chose different loss functions for training from the perspectives of global anomaly detection and local anomaly localization.

\subsubsection{Global Loss}
We employ cross-entropy loss to optimize our global anomaly score. Cross-entropy loss is a commonly used loss function in various tasks and its formula is as follows:
\begin{equation}
    L_{ce} = -\sum_{i=1}^{n}{y_i log(p_i)},\label{eq:ce}
\end{equation}
where $n$ is the number of instances,  $y_i$ is the true label for instance $i$ and $p_i$ is the score for instance $i$. we use cross-entropy loss to calculate our global loss:
\begin{equation}
    L_{global} = L_{ce}(S_{global}, Label),
\end{equation}

\noindent where $S_{global}$ represents the global anomaly score calculated in Sec~3.2.2, and $Label$ denotes the label indicating whether the image is anomalous or not.

\subsubsection{Local Loss}
We employ Focal loss~\cite{lin2017focal} and Dice loss~\cite{milletari2016v} to optimize our anomaly map $M$. Focal Loss and Dice Loss are common loss functions used in semantic segmentation tasks. Specifically, Focal Loss is particularly effective in addressing class imbalance issues, making it well-suited for anomaly localization tasks where the proportion of anomaly regions is relatively small. Focal loss can be calculated by Eq.~(\ref{eq:focal}):
\begin{equation}
    L_{f} = -\frac{1}{n}\sum_{i=1}^{n}{(1-p_i)^{\gamma}log(p_i)},\label{eq:focal}
\end{equation}
where $n = H \times W$ represents the total number of pixels, $p_i$ is the predicted probability of the positive classes and $\gamma$ is a tunable parameter for adjusting the weight of hard-to-classify samples. In our implementation, we set $\gamma$ to 2.

Dice loss can be calculated by Eq.~(\ref{eq:dice}):
\begin{equation}
    L_{d} = -\frac{\sum_{i=1}^{n}{y_i\hat{y}_i}}{\sum_{i=1}^{n}{y_i^2}+\sum_{i=1}^{n}{\hat{y}^2_i}},\label{eq:dice}
\end{equation}
where $n = H \times W$, $y_i$ is the output of decoder and $\hat{y}_i$ is the ground truth value. 

Our local loss can be calculated by Eq.~(\ref{eq:local}):

\begin{equation}
    L_{local} = L_{f}(M^a, G) + L_{d}(M^a, G) + L_{d}(M^n, 1 - G),\label{eq:local}
\end{equation}
where G denotes the ground truth.

% \subsubsection{Final Loss}
% With a hyperparameter $\beta$ to balance the global and local losses, our final loss can be calculated by Eq.~(\ref{eq:final_loss}):
% \begin{equation}
%     L_{total} = L_{global} + \beta L_{local},\label{eq:final_loss}
% \end{equation}

% \noindent where $\beta$ is set to 1 in our experiments.

%% file: sec/4_experiment.tex
\section{Experiments}

\begin{table*}[]
\centering
\small
\begin{tabular}{@{}ccccccc@{}}
\toprule
\multirow{2}{*}{Method} &
  \multirow{2}{*}{Backbone} &
  \multirow{2}{*}{Anomaly Description} &
  \multicolumn{2}{c}{VisA} &
  \multicolumn{2}{c}{MVTec-AD} \\ \cmidrule(l){4-7} 
                                                             &                &                    & Image-AUC & Pixel-AUC & Image-AUC & Pixel-AUC \\ \midrule
CLIP~\cite{radford2021learning}        & ViT-L/14@336px & normal / anomalous & 66.4      & 46.6      & 74.1      & 38.4      \\
CLIP-AC~\cite{radford2021learning}     & ViT-L/14@336px & normal / anomalous & 65.0      & 47.8      & 71.5      & 38.2      \\
WinCLIP~\cite{jeong2023winclip} &
  ViT-B/16@240px &
  state ensemble &
  78.1 &
  79.6 &
  \textbf{91.8} &
  85.1 \\
APRIL-GAN~\cite{chen2023zero}          & ViT-L/14@336px & state ensemble     & 78.0      & 94.2      & 86.1      & 87.6      \\
AnomalyCLIP~\cite{zhou2023anomalyclip} & ViT-L/14@336px & normal / damaged   & 82.1      & 95.5      & 91.5      & 91.1      \\
AnomalyCLIP-                                                 & ViT-L/14@336px & normal / damaged   & 81.7      & 95.0      & 90.8      & 89.5      \\
\textbf{FiLo~(ours)} &
  ViT-L/14@336px &
  fine-grained description &
  \textbf{83.9} &
  \textbf{95.9} &
  91.2 &
  \textbf{92.3} \\ \bottomrule
\end{tabular}
\caption{Comparison results between FiLo and other ZSAD methods. The best-performing method is in \textbf{bold}.}
\label{tab:main-results}
\end{table*}

\begin{table}[]
\centering
\small
\begin{tabular}{@{}ccc@{}}
\toprule
Setup                      & VisA         & MVTec        \\ \midrule
CLIP baseline              & (65.0, 47.8) & (71.5, 38.2) \\
+ generic {[}state{]}      & (65.4, 83.9) & (79.9, 83.5) \\
\textbf{+ fine-grained {[}anomaly class{]}} & \textbf{(71.2, 85.5)} & \textbf{(80.8, 83.8)} \\ \bottomrule
\end{tabular}
\caption{Ablation results of anomaly descriptions. Results are displayed in the format of~(Image-AUC, Pixel-AUC).}
\label{tab:ab-fg-des}
\end{table}

\begin{table}[]
\centering
\begin{tabular}{@{}ccc@{}}
\toprule
Setup                      & VisA         & MVTec        \\ \midrule
CLIP baseline              & (65.0, 47.8) & (71.5, 38.2) \\
+ learnable template       & (72.5, 93.1) & (82.1, \textbf{85.2}) \\
+ \textbf{fine-grained description} & \textbf{(78.1, 93.2)} & \textbf{(85.8,} 85.1) \\ \bottomrule
\end{tabular}
\caption{Ablation results of text template. Results are displayed in the format of~(Image-AUC, Pixel-AUC).}
\label{tab:ab-coop}
\end{table}

% \usepackage{multirow}
\begin{table*}[]
\centering
\small
\begin{tabular}{@{}cccccccc@{}}
\toprule
\multirow{2}{*}{Grounding} &
  \multirow{2}{*}{Position Enhancement} &
  \multicolumn{2}{c}{MMCI} &
  \multicolumn{2}{c}{VisA} &
  \multicolumn{2}{c}{MVTec} \\ \cmidrule(l){5-8} 
             &              & Multi-shape  & Multi-scale  & Image-AUC & Pixel-AUC & Image-AUC & Pixel-AUC \\ \midrule
             &              &              &              & 78.1      & 93.2      & 85.8      & 85.1      \\
$\checkmark$ &              &              &              & 78.1      & 93.4      & 85.8      & 85.3      \\
$\checkmark$ & $\checkmark$ &              &              & 78.6      & 93.6      & 85.5      & 85.7      \\
$\checkmark$ & $\checkmark$ & $\checkmark$ &              & 79.2      & 95.3      & 86.2      & 89.4      \\
$\checkmark$ & $\checkmark$ &              & $\checkmark$ & 80.7      & 95.6      & 88.9      & 91.4      \\
$\checkmark$ &
  $\checkmark$ &
  $\checkmark$ &
  $\checkmark$ &
  \textbf{83.9} &
  \textbf{95.9} &
  \textbf{91.2} &
  \textbf{92.3} \\ \bottomrule
\end{tabular}
\caption{The results of ablation experiments for each proposed modules in HQ-Loc.}
\label{tab:ab-hq-loc}
\end{table*}

\subsection{Datasets}
Our experiments primarily focus on two datasets: MVTec~\cite{bergmann2019mvtec} and VisA~\cite{zou2022spot}. MVTec~\cite{bergmann2019mvtec} is one of the most widely used industrial anomaly detection datasets, containing 5354 images of both normal and abnormal samples from 15 different object categories, with resolutions ranging from $700 \times 700$ to $1024\times 1024$ pixels. VisA~\cite{zou2022spot} is an emerging industrial anomaly detection dataset comprising 10821 images of normal and abnormal samples covering 12 image categories, with resolutions around $1500 \times 1000$ pixels. Similar to APRIL-GAN~\cite{chen2023zero} and AnomalyCLIP~\cite{zhou2023anomalyclip}, we conduct supervised training on the test set of one dataset and directly performed zero-shot testing on the other dataset.

\subsection{Evaluation Metrics}
Following existing AD methods~\cite{you2022unified,defard2021padim}, we employ the Area Under the receiver operating Characteristic~(AUC) as our evaluation metric, with image-level and pixel-level AUC used to assess anomaly detection and anomaly localization performance, respectively. 
% In addition, existing methods~\cite{zhou2023anomalyclip,roth2022towards, bergmann2020uninformed} have demonstrated that solely using the AUC metric is insufficient to fully evaluate the performance of anomaly detection and localization. Therefore, we also employed the Average Precision~(AP) for anomaly detection and the AUPRO~\cite{bergmann2020uninformed} metric for anomaly segmentation to further assess the performance of our method.

\subsection{Implementation Details}
We utilize the publicly available CLIP-L/14@336px model as our backbone, with frozen parameters for CLIP's Text Encoder and Image Encoder. Training is conducted on either the MVTec or VisA dataset, with zero-shot testing performed on the other dataset. For intermediate-level patch-based image features, we employ features from the 6-th, 12-th, 18-th, and 24-th layers of the CLIP Image Encoder. Starting from the 6-th layer, both QKV Attention and V-V Attention results are simultaneously utilized, where the outputs of QKV Attention are aligned with text features through a simple linear layer, and the outputs of V-V Attention are inputted into the MMCI module for multi-scale, multi-shape deep interaction with text features. During training, input images are resized to a resolution of $518 \times 518$, and the AdamW~\cite{loshchilov2017decoupled} optimizer is used to optimize model parameters for 15 epochs. The learning rate for learnable text vectors is set to 1e-3, while the learning rate for the MMCI module is set to 1e-4. After that, we train the adapter for 5 epochs with a learning rate of 1e-5. Additionally, due to the varying number of fine-grained anomaly descriptions for each item category, training is conducted with a batch size of 1. Following previous methods~\cite{you2022unified, zhou2023anomalyclip}, a Gaussian filter with $\sigma=4$ is applied to obtain a smoother anomaly score map during testing.

\subsection{Main Results}
To demonstrate the effectiveness of our FiLo, we compare FiLo with several existing ZSAD methods, including CLIP~\cite{radford2021learning}, CLIP-AC~\cite{radford2021learning}, WinCLIP~\cite{jeong2023winclip}, APRIL-GAN~\cite{chen2023zero}, and AnomalyCLIP~\cite{zhou2023anomalyclip}. Following~\cite{zhou2023anomalyclip}, for CLIP, we conduct experiments using simple text prompts \textit{A photo of a normal [class].} and \textit{A photo of an anomalous [class]}, and we add more text prompt templates that are recommended for ImageNet dataset for CLIP-AC. Results for WinCLIP~\cite{jeong2023winclip}, APRIL-GAN~\cite{chen2023zero}, and AnomalyCLIP~\cite{zhou2023anomalyclip} are adopted from their respective papers. Specifically, AnomalyCLIP~\cite{zhou2023anomalyclip} incorporates additional learnable embeddings in the CLIP Text Encoder, while other methods, including our FiLo, directly utlize the frozen parameters of CLIP. To ensure fair comparison, we reproduce AnomalyCLIP without learnable embeddings, which is referred as AnomalyCLIP-. 

Table~\ref{tab:main-results} presents the experimental results of FiLo and existing methods on the VisA and MVTec datasets, which demonstrates superiority of FiLo across most metrics on both datasets, validating the effectiveness of our FG-Des and HQ-Loc modules. Compared to the state-of-the-art ZSAD method AnomalyCLIP~\cite{zhou2023anomalyclip}, after introducing the FG-Des and HQ-Loc modules, FiLo achieves a 1.1\% improvement in image-level AUC and a 0.4\% improvement in pixel-level AUC on the VisA dataset. Additionally, FiLo also achieves a 1.2\% improvement in pixel-level AUC on the MVTec dataset.

% \begin{table*}[]
% \centering
% \begin{tabular}{@{}ccccc@{}}
% \toprule
% \multirow{2}{*}{Method} & \multicolumn{2}{c}{MVTec-AD}                     & \multicolumn{2}{c}{VisA}                         \\ \cmidrule(l){2-5} 
%                         & (Image-AUC, Image-AP) & (Pixel-AUC, Pixel-AUPRO) & (Image-AUC, Image-AP) & (Pixel-AUC, Pixel-AUPRO) \\ \midrule
% CLIP~\cite{radford2021learning}         & (74.1, 87.6) & (38.4, 11.3) & (66.4, 71.5) & (46.6, 14.8) \\
% CLIP-AC~\cite{radford2021learning}      & (71.5, 86.4) & (38.2, 11.6) & (65.0, 70.1) & (47.8, 17.3) \\
% WinCLIP~\cite{jeong2023winclip}      & \textbf{(91.8, 96.5)} & (85.1, 64.6) & (78.1, 81.2) & (79.6, 56.8) \\
% APRIL-GAN~\cite{chen2023zero}    & (86.1, 93.5) & (87.6, 44.0) & (78.0, 81.4) & (94.2, 86.8) \\
% AnomalyCLIP~\cite{zhou2023anomalyclip} & (91.5, 96.2) & (91.1, 81.4) & (82.1, 85.4) & (95.5, 87.0) \\
% AnomalyCLIP- & (90.8, 96.0) & (89.5, 81.2) & (81.7, 85.2) & (95.0, 85.3) \\
% \textbf{FiLo(ours)}   & (91.2, -)    & \textbf{(92.1, -)}    & \textbf{(83.9, 87.3)}    & \textbf{(95.9, 85.1)}    \\ \bottomrule
% \end{tabular}
% \caption{Comparison results between our method and SOTA ZSAD methods. The best-performing method is in bold.}
% \label{tab:main-results}
% \end{table*}

\begin{figure*}[t]
  \centering
  \includegraphics[width=0.99\textwidth]{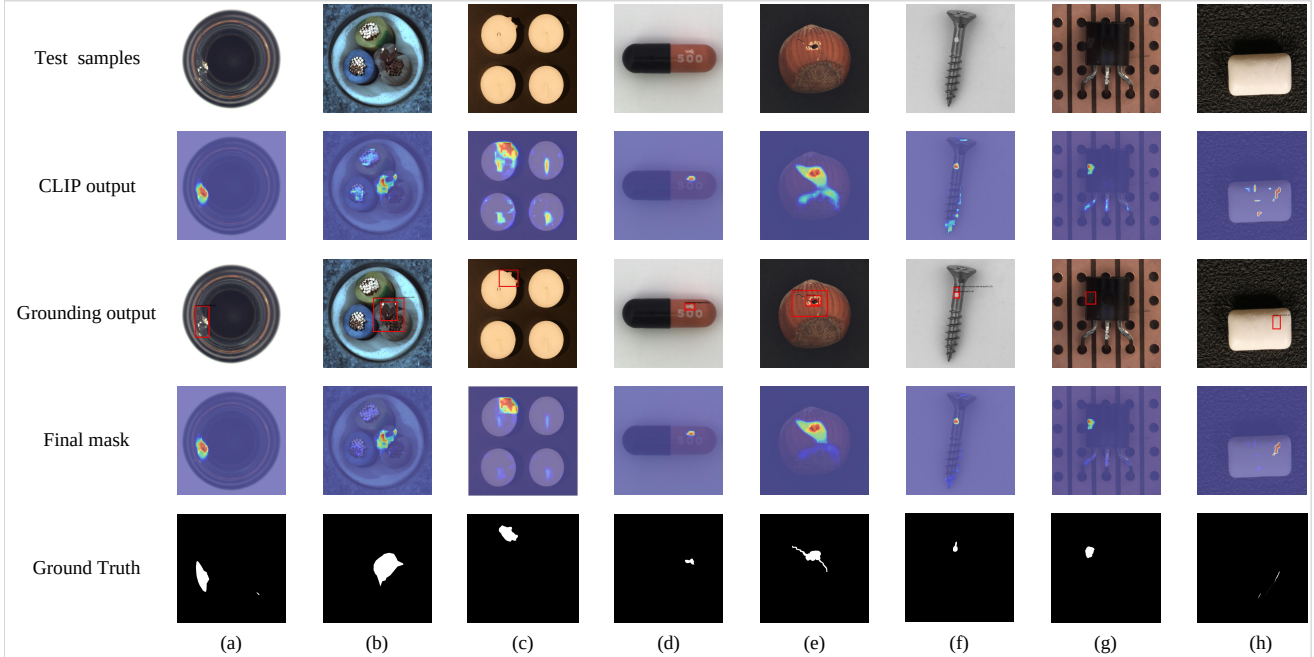} 
  \caption{Visualization result of FiLo on MVTec and VisA datasets. "CLIP output" refers to the localization results without HQ-Loc, while "Final mask" represents the final localization result.}
  \label{fig:visual}
\end{figure*}

\subsection{Ablation Study}
To investigate the effectiveness of each proposed module, we conduct extensive ablation experiments on the VisA and MVTec datasets, confirming the efficacy of every component in our approach, including fine-grained descriptions, learnable text templates, grounding, position enhancement and MMCI. Table~\ref{tab:ab-fg-des}, Table~\ref{tab:ab-coop} and Table~\ref{tab:ab-hq-loc} present the experimental results of FiLo on the MVTec and VisA datasets.

% Specifically, adaptively learned text templates and the fine-grained anomaly descriptions in FG-Des, as well as the effectiveness of each module in HQ-Loc, including the preliminary localization via Grounding, the location-enhanced text prompt, and MMCI are all validated.

In Table~\ref{tab:ab-fg-des}, we initially employ the same setup as CLIP-AC as our baseline, using simple two-category texts \textit{A photo of a normal [class]} and \textit{A photo of an anomalous [class]}. Upon realizing that the simple words "normal" and "anomalous" alone did not effectively distinguish between normal and abnormal samples, we modify the sentence structure to \textit{A photo of a [state] [class]}, where [state] encompasses some generic descriptions for normal (e.g., perfect, flawless) and abnormal (e.g., damaged, defective) states, and observe a significant performance improvement with the introduction of more detailed [state] descriptions. Subsequently, we utilize LLMs to generate more fine-grained [anomaly class] for each class of items, resulting in further performance enhancements. This experiment underscores the effectiveness of fine-grained anomaly descriptions.

In Table~\ref{tab:ab-coop}, also starting from the CLIP baseline, we first replace all parts of the text except for [class] with learnable vectors, i.e., [v1][v2]...[vn][class]. We find that compared to handcrafted text, the text vectors learned by the model are more suitable for anomaly detection tasks, exhibiting higher detection and localization accuracy. Further, by combining the learned text vectors with detailed anomaly descriptions generated by LLMs as described earlier, we utilize the text prompt [v1][v2]...[vn][state][class] with [anomaly class], resulting in significant improvements.

In Table~\ref{tab:ab-hq-loc}, we experiment with each component of HQ-Loc. From the table, it can be observed that both Grounding and Position Enhancement contribute to improvements in pixel-level AUC. Additionally, the MMCI module, which integrates multi-shape and multi-size capabilities, can effectively detect anomalies of various sizes and shapes, resulting in performance enhancements in both detection and localization aspects.

% In the ablation experiments, we employed a simple binary text template \textit{A photo of a [state] [class name]} as the baseline, where [state] encompassed some generic descriptions for normal (e.g., perfect, flawless) and abnormal (e.g., damaged, defective) states. We then utilized a basic linear layers to align intermediate patch features with textual features, thereby enabling the baseline method to possess anomaly localization capabilities.

% Initially, by augmenting the baseline with fine-grained anomaly descriptions and adaptively learned text prompts, we achieved a more detailed and flexible textual representation, leading to a noticeable improvement in both image-level and pixel-level AUC. Furthermore, the incorporation of grounding for preliminary localization and position enhancement significantly reduced false positives in the background regions, resulting in a further enhancement in pixel-level AUC. Finally, the utilization of the MMCI module enabled the comprehensive detection of anomalies of different sizes and shapes, leading to additional improvements in both image-level and pixel-level AUC, culminating in the final FiLo method.

\subsection{Visulization Results}

Figure~\ref{fig:visual} illustrates the visualization results of FiLo on the MVTec and VisA datasets. In the absence of any prior access to data from the target dataset, FiLo achieves anomaly localization results that closely resemble the ground truth, showcasing FiLo's robust ZSAD capability.

As observed in the second row of Figure~\ref{fig:visual}, directly computing the similarity between all patch features extracted using CLIP and textual features representing normal and abnormal semantics often yields imprecise anomaly localization results. This approach sometimes leads to false positives in non-anomalous objects or background regions of the image. However, by employing HQ-Loc's grounding for preliminary localization and position enhancement, the final output effectively mitigates this phenomenon.

Furthermore, during the preliminary localization process, Grounding associates each bounding box with matched textual descriptions, indicating the type of anomaly present in that area. For instance, in Figure~\ref{fig:visual}(e), the corresponding text for the bounding box accurately identifies anomalies on the hazelnut: "hole" and "crack".

%% file: sec/5_conclusion.tex
\section{Conclusion}

Our FiLo method represents a significant advancement in the field of Zero-Shot Anomaly Detection (ZSAD), effectively addressing prevalent challenges in both anomaly detection and localization. Our FG-Des method harnesses the capabilities of Large Language Models (LLMs) by generating specific descriptions for potential anomaly types associated with each object category. This approach notably enhances both the precision and interpretability of anomaly detection. Furthermore, our devised HQ-Loc strategy effectively mitigates the deficiencies of existing methods in terms of anomaly localization accuracy, particularly demonstrating superior performance in localizing anomalies of various sizes and shapes. Extensive experiments validate the superiority of FiLo across multiple datasets, affirming its efficacy and practicality in the realm of zero-shot anomaly detection tasks.

%% file: sec/6_suppl.tex
\clearpage
\maketitlesupplementary

\appendix

% % \onecolumn

% \begin{center}
%     {\bf {\LARGE Supplementary Material}}
% \end{center}
% \begin{center}
% {\bf {\large FiLo: Zero-Shot Anomaly Detection by Fine-Grained Description and High-Quality Localization \\[20pt]}}
% \end{center}

\section{Fine-grained ZSAD performance}

In the main paper, we have compared FiLo with existing ZSAD methods on anomaly detection and localization across the MVTec~\cite{bergmann2019mvtec} and VisA~\cite{zou2022spot} datasets. Our evaluation primarily utilizes Image-level AUC and Pixel-level AUC as metrics for detection and localization, respectively. Here, we provide detailed performance analysis of FiLo and other ZSAD methods at the fine-grained data subset level, including the methods we using for comparison: CLIP~\cite{radford2021learning}, CLIP-AC~\cite{radford2021learning}, WinCLIP~\cite{jeong2023winclip}, APRIL-GAN~\cite{chen2023zero} and AnomalyCLIP~\cite{zhou2023anomalyclip}.

Tables~\ref{tab:seg-mvtec} and Tables~\ref{tab:seg-visa} depict the anomaly localization performance of FiLo on the MVTec and VisA datasets, and the anomaly detection performance of FiLo on the VisA and MVTec datasets is showcased in Table~\ref{tab:cls-visa} and Table~\ref{tab:cls-mvtec} respectively. Across the 15 classes in the MVTec dataset, FiLo achieves the highest Pixel-level AUC in 12 classes, while in the VisA dataset comprising 12 classes, FiLo attains the highest Pixel-level AUC in 8 classes. Notably, FiLo surpasses the state-of-the-art method AnomalyCLIP~\cite{zhou2023anomalyclip} by 1.1\% on Pixel-level AUC in the MVTec dataset and by 0.4\% in the VisA dataset, demonstrating the efficacy of FiLo.

\begin{table*}[]
\centering

\begin{tabular}{@{}ccccccc@{}}
\toprule
Object name & CLIP~\cite{radford2021learning} & CLIP-AC~\cite{radford2021learning} & WinCLIP~\cite{jeong2023winclip} & APRIL-GAN~\cite{chen2023zero} & AnomalyCLIP~\cite{zhou2023anomalyclip}   & \textbf{FiLo~(ours)} \\ \midrule
Carpet      & 11.5 & 10.7    & 95.4    & 98.4 & \underline{98.8}        & \textbf{99.4} \\
Bottle      & 17.5 & 23.3    & 89.5    & 83.4 & \underline{90.4}        & \textbf{92.6} \\
Hazelnut    & 25.2 & 34.0    & 94.3    & 96.1 & \underline{97.1}        & \textbf{97.6} \\
Leather     & 9.9  & 5.6     & 96.7    & \underline{99.1} & 98.6        & \textbf{99.4} \\
Cable       & 37.4 & 37.5    & 77.0    & 72.3 & \textbf{78.9}        & \underline{78.4} \\
Capsule     & 50.9 & 49.1    & 86.9    & 92.0 & \underline{95.8}        & \textbf{96.9} \\
Grid        & 8.7  & 11.9    & 82.2    & 95.8 & \underline{97.3}       & \textbf{97.8} \\
Pill        & 55.8 & 60.8    & 80.0    & 76.2 & \textbf{92}          & \underline{89.1} \\
Transistor  & 51.1 & 48.5    & \underline{74.7}    & 62.4 & 71          & \textbf{74.8} \\
Metal\_nut  & 43.9 & 53.6    & 61.0    & 65.4 & \textbf{74.4}        & \underline{72.5} \\
Screw       & 80.1 & 76.4    & 89.6    & \underline{97.8} & 97.5        & \textbf{98.1} \\
Toothbrush  & 36.3 & 35.0    & 86.9    & \underline{95.8} & 91.9        & \textbf{96.0}   \\
Zipper      & 51.5 & 44.7    & \underline{91.6}    & 91.1 & 91.4        & \textbf{96.6} \\
Tile        & 49.9 & 39.1    & 77.6    & 92.7 & \underline{94.6}        & \textbf{97.4} \\
Wood        & 45.7 & 42.4    & 93.4    & 95.8 & \underline{96.5}        & \textbf{98.3} \\ \midrule
Mean        & 38.4 & 38.2    & 85.1    & 87.6 & \underline{91.1}        & \textbf{92.3} \\ \bottomrule
\end{tabular}
\caption{Fine-grained data-subset-wise performance comparison (AUROC) for anomaly localization on  MVTec-AD. The best performance is
in \textbf{bold}, and the second-best is \underline{underlined}.}
\label{tab:seg-mvtec}
\end{table*}

\begin{table*}[]
\centering

\begin{tabular}{@{}ccccccc@{}}
\toprule
Object name & CLIP~\cite{radford2021learning} & CLIP-AC~\cite{radford2021learning} & WinCLIP~\cite{jeong2023winclip} & APRIL-GAN~\cite{chen2023zero} & AnomalyCLIP~\cite{zhou2023anomalyclip}   & \textbf{FiLo~(ours)}          \\ \midrule
Candle                                & 33.6 & 50.0    & 88.9    & 97.8 & \textbf{98.8} & \underline{98.7}          \\
Capsules                              & 56.8 & 61.5    & 81.6    & \textbf{97.5} & \underline{95.0} & 92.3          \\
Cashew                                & 64.5 & 62.5    & 84.7    & 86.0 & \underline{93.8}          & \textbf{95.1} \\
Chewinggum                            & 43.0 & 56.5    & 93.3    & \textbf{99.5} & 99.3          & \underline{99.4} \\
Fryum                                 & 45.6 & 62.7    & 88.5    & 92.0 & \underline{94.6}          & \textbf{95.2} \\
Macaroni1                             & 20.3 & 22.9    & 70.9    & \underline{98.8} & 98.3          & \textbf{99.1} \\
Macaroni2                             & 37.7 & 28.8    & 59.3    & \underline{97.8} & 97.6          & \textbf{98.1} \\
Pcb1                                  & 57.8 & 51.6    & 61.2    & 92.7 & \underline{94.1}          & \textbf{94.4} \\
Pcb2                                  & 34.7 & 38.4    & 71.6    & 89.7 & \underline{92.4}          & \textbf{93.7} \\
Pcb3                                  & 54.6 & 44.6    & 85.3    & \underline{88.4} & \underline{88.4}          & \textbf{90.8} \\
Pcb4                                  & 52.1 & 49.9    & 94.4    & 94.6 & \underline{95.7}          & \textbf{95.8} \\
Pipe\_fryum                           & 58.7 & 44.7    & 75.4    & 96.0 & \textbf{98.2} & \underline{97.7}          \\ \midrule
Mean                                  & 46.6 & 47.8    & 79.6    & 94.2 & \underline{95.5}          & \textbf{95.9} \\ \bottomrule
\end{tabular}
\caption{Fine-grained data-subset-wise performance comparison (AUROC) for anomaly localization on VisA. The best performance is
in \textbf{bold}, and the second-best is \underline{underlined}.}
\label{tab:seg-visa}
\end{table*}

\begin{figure*}[h]
  \centering
  \includegraphics[width=0.8\textwidth]{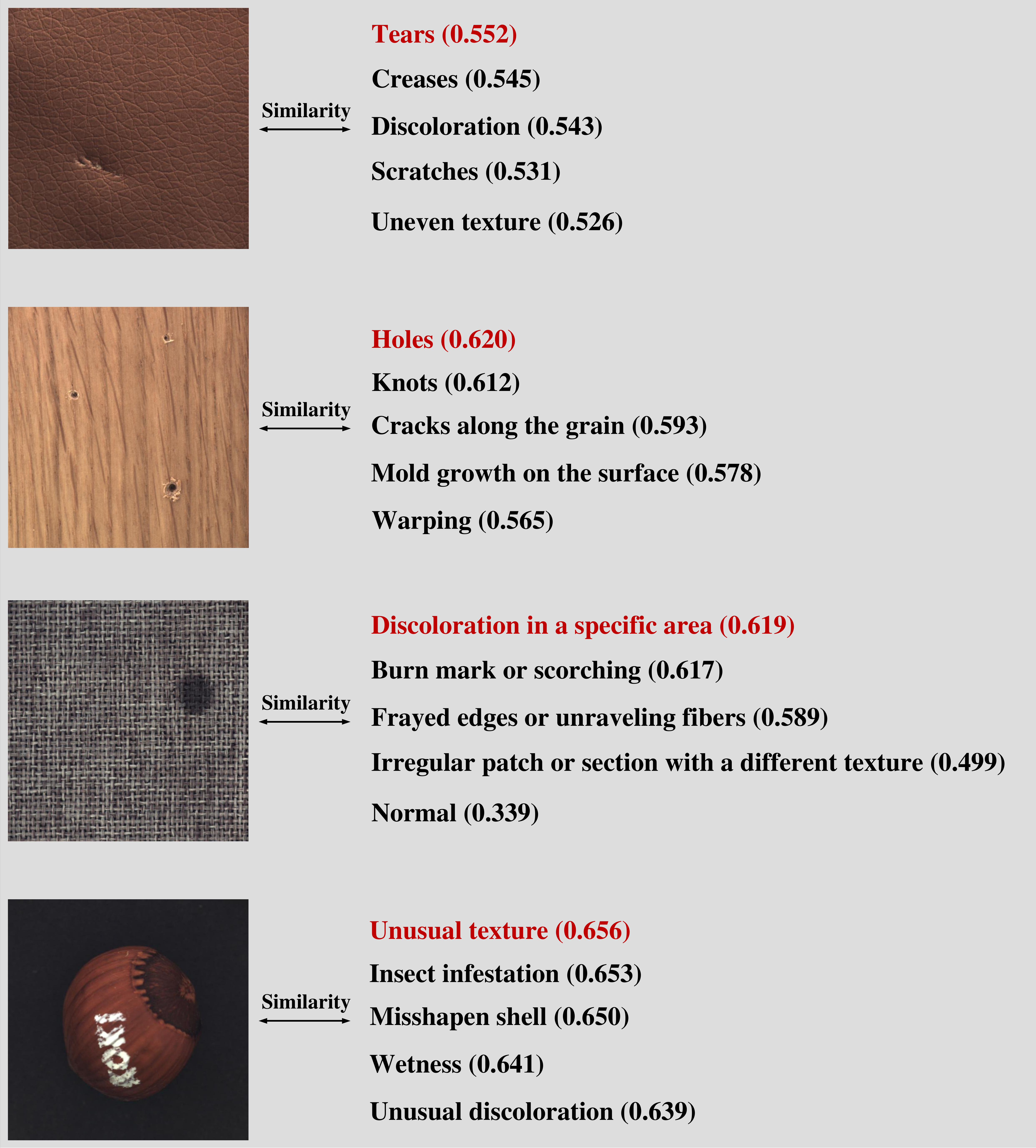} 
  \caption{Illustration of similarities between images and different fine-grained anomaly descriptions.}
  \label{fig:detail_match}
\end{figure*}

\section{Fine-Grained Anomaly Descriptions}

Table~\ref{tab:detail-mvtec} and Table~\ref{tab:detail-visa} present the detailed anomaly types generated by leveraging LLM for each category within the MVTec and VisA datasets. During the inference process with FiLo, we substitute these detailed anomaly descriptions generated by LLM for the "[ANOMALY CLASS]" portion in the text template to obtain the detailed anomaly description content for each category of items.

\begin{table*}[]
\centering

\begin{tabular}{@{}ccccccc@{}}
\toprule
Object name & CLIP~\cite{radford2021learning} & CLIP-AC~\cite{radford2021learning} & WinCLIP~\cite{jeong2023winclip} & APRIL-GAN~\cite{chen2023zero} & AnomalyCLIP~\cite{zhou2023anomalyclip}   & \textbf{FiLo~(ours)} \\ \midrule
Carpet               & 96   & 93.1    & \textbf{100.0}   & 99.5 & \textbf{100.0} & \underline{99.9}                 \\
Bottle               & 45.9 & 46.1    & \textbf{99.2}    & 92.0 & 89.3           & \underline{98.6}        \\
Hazelnut             & 88.7 & 91.1    & \underline{93.9}    & 89.6 & \textbf{97.2}  & 92.8                 \\
Leather              & 99.4 & 99.5    & \textbf{100.0}   & 99.7 & \underline{99.8}           & \textbf{100}         \\
Cable                & 58.1 & 46.6    & \underline{86.5}    & \textbf{88.4} & 69.8           & 77.9        \\
Capsule              & 71.4 & 68.8    & 72.9    & 79.9 & \textbf{89.9}  & \underline{89.2}                 \\
Grid                 & 72.5 & 63.7    & \textbf{98.8}    & 86.3 & 97.0           & \underline{97.4}        \\
Pill                 & 73.6 & 73.8    & 79.1    & 80.5 & \underline{81.8}           & \textbf{87.8}        \\
Transistor           & 48.8 & 51.2    & \underline{88.0}    & 80.8 & \textbf{92.8}  & 80.5                 \\
Metal\_nut           & 62.8 & 63.4    & \textbf{97.1}    & 68.4 & \underline{93.6}  & 77                   \\
Screw                & 78.2 & 66.7    & \underline{83.3}    & \textbf{84.9} & 81.1  & 74.5                 \\
Toothbrush           & 73.3 & \underline{89.2}    & 88.0    & 53.8 & 84.7           & \textbf{94.2}        \\
Zipper               & 60.1 & 36.1    & 91.5    & 89.6 & \textbf{98.5}  & \underline{98.1}                 \\
Tile                 & 88.5 & 89.0    & \textbf{100.0}   & \underline{99.9} & \textbf{100.0} & \textbf{100}         \\
Wood                 & 94   & 94.9    & \underline{99.4}    & 99.0 & 96.8           & \textbf{99.7}        \\ \midrule
Mean                 & 74.1 & 71.5    & \textbf{91.8}    & 86.1 & \underline{91.5}  & 91.2                 \\ \bottomrule
\end{tabular}
\caption{Fine-grained data-subset-wise performance comparison (AUROC) for anomaly detection on MVTec AD. The best performance is
in \textbf{bold}, and the second-best is \underline{underlined}.}
\label{tab:cls-mvtec}
\end{table*}

\begin{table*}[]
\centering

\begin{tabular}{@{}ccccccc@{}}
\toprule
Object name & CLIP~\cite{radford2021learning} & CLIP-AC~\cite{radford2021learning} & WinCLIP~\cite{jeong2023winclip} & APRIL-GAN~\cite{chen2023zero} & AnomalyCLIP~\cite{zhou2023anomalyclip}   & \textbf{FiLo~(ours)} \\ \midrule
Candle               & 37.9 & 33.0    & \textbf{95.4}    & \underline{83.8} & 79.3 & 79.3        \\
Capsules             & 69.7 & 75.3    & \textbf{85.0}    & 61.2 & \textbf{81.5} & 80.9                 \\
Cashew               & 69.1 & 72.7    & \textbf{92.1}    & 87.3 & 76.3          & \underline{90}          \\
Chewinggum           & 77.5 & 76.9    & 96.5    & 96.4 & \underline{97.4}          & \textbf{98.4}        \\
Fryum                & 67.2 & 60.9    & 80.3    & \textbf{94.3} & \underline{93.0} & 88.3                 \\
Macaroni1            & 64.4 & 67.4    & 76.2    & 71.6 & \underline{87.2}          & \textbf{88.3}        \\
Macaroni2            & 65   & 65.7    & 63.7    & 64.6 & \textbf{73.4} & \underline{68.5}                 \\
Pcb1                 & 54.9 & 43.9    & 73.6    & 53.4 & \underline{85.4}          & \textbf{87}          \\
Pcb2                 & 62.6 & 59.5    & 51.2    & \underline{71.8} & 62.2          & \textbf{77.6}        \\
Pcb3                 & 52.2 & 49.0    & \textbf{73.4}    & 66.8 & 62.7          & \underline{69.5}        \\
Pcb4                 & 87.7 & 89.0    & 79.6    & \underline{95.0} & 93.9          & \textbf{95.7}        \\
Pipe\_fryum          & 88.8 & 86.4    & 69.7    & \underline{89.9} & \textbf{92.4} & 83.8                 \\ \midrule
Mean                 & 66.4 & 65.0    & 78.1    & 78.0 & \underline{82.1}          & \textbf{83.9}        \\ \bottomrule
\end{tabular}
\caption{Fine-grained data-subset-wise performance comparison (AUROC) for anomaly detection on VisA. The best performance is
in \textbf{bold}, and the second-best is \underline{underlined}.}
\label{tab:cls-visa}
\end{table*}

\begin{table*}[]
\centering

\begin{tabular}{@{}cc@{}}
\toprule
Class name & Descriptions \\ \midrule
Carpet &
  \begin{tabular}[c]{@{}c@{}}discoloration in a specific area, irregular patch or section with a different texture, \\ frayed edges or unraveling fibers, burn mark or scorching\end{tabular} \\ \midrule
Grid &
  \begin{tabular}[c]{@{}c@{}}crooked, cracks, excessive gaps, discoloration, deformation, missing, \\ inconsistent spacing between grid elements, corrosion, visible signs, chipping\end{tabular} \\ \midrule
Leather &
  \begin{tabular}[c]{@{}c@{}}scratches, discoloration, creases, uneven texture, tears, \\ brittleness, damage, seams, heat damage, mold\end{tabular} \\ \midrule
Tile &
  \begin{tabular}[c]{@{}c@{}}chipped, irregularities, discoloration, efflorescence, warping, \\ missing, depressions, lippage, fungus, damage\end{tabular} \\ \midrule
Wood &
  \begin{tabular}[c]{@{}c@{}}knots, warping, cracks along the grain, mold growth on the surface, staining from water damage, \\ wood rot, woodworm holes, rough patches, protruding knots\end{tabular} \\ \midrule
Bottle &
  \begin{tabular}[c]{@{}c@{}}cracked large, cracked small, dented large, dented small, leaking, discolored, \\ deformed, missing cap, excessive condensation, unusual odor\end{tabular} \\ \midrule
Cable &
  \begin{tabular}[c]{@{}c@{}}twisted, knotted cable strands, detached connectors, excessive stretching, \\ dents, corrosion, scorching along the cable, exposed conductive material\end{tabular} \\ \midrule
Capsule &
  \begin{tabular}[c]{@{}c@{}}irregular shape, discoloration coloring, crinkled, uneven seam, \\ condensation inside the capsule, foreign particles, unusually soft or hard\end{tabular} \\ \midrule
Hazelnut &
  \begin{tabular}[c]{@{}c@{}}fungal growth, Unusual discoloration, rotten or foul odor emanating, insect infestation, \\ wetness, misshapen shell, unusually thin, contaminants, unusual texture\end{tabular} \\ \midrule
Metal\_nut &
  \begin{tabular}[c]{@{}c@{}}cracks, irregular threading, corrosion, missing, distortion, signs of discoloration, \\ excessive wear on contact surfaces, inconsistent texture\end{tabular} \\ \midrule
Pill &
  \begin{tabular}[c]{@{}c@{}}irregular shape, crumbling texture, excessive powder, uneven coating, \\ presence of air bubbles, disintegration, abnormal specks\end{tabular} \\ \midrule
Screw &
  \begin{tabular}[c]{@{}c@{}}rust on the surface, bent, damaged threads, stripped threads, deformed top, \\ coating damage, uneven grooves, inconsistent size\end{tabular} \\ \midrule
Toothbrush &
  \begin{tabular}[c]{@{}c@{}}loose bristles, uneven bristle distribution, excessive shedding of bristles, \\ staining on the bristles, abrasive texture, irregularities in the shape\end{tabular} \\ \midrule
Transistor &
  \begin{tabular}[c]{@{}c@{}}burn marks, detached leads, signs of corrosion, irregularities in the shape, \\ presence of cracks or fractures, signs of physical trauma, irregularities in the surface texture\end{tabular} \\ \midrule
Zipper &
  bent, frayed, misaligned, excessive stiffness, corroded, detaches, loose, warped \\ \bottomrule
\end{tabular}
\caption{Fine-Grained anomaly description of every object within MVTec dataset.}
\label{tab:detail-mvtec}
\end{table*}

\begin{table*}[]
\centering

\begin{tabular}{@{}cc@{}}
\toprule
Class name &
  Descriptions \\ \midrule
Candle &
  \begin{tabular}[c]{@{}c@{}}cracks or fissures in the wax, Wax pooling unevenly around the wick, tunneling, incomplete wax melt pool, \\ irregular or flickering flame, other, extra wax in candle, wax melded out of the candle\end{tabular} \\ \midrule
Capsules &
  \begin{tabular}[c]{@{}c@{}}uneven capsule size, capsule shell appears brittle, excessively soft, \\ dents, condensation, irregular seams or joints, specks\end{tabular} \\ \midrule
Cashew &
  \begin{tabular}[c]{@{}c@{}}uneven coloring, fungal growth, presence of foreign objects, \\ unusual texture, empty shells, signs of moisture, stuck together\end{tabular} \\ \midrule
Chewinggum &
  consistency, presence of foreign objects, uneven coloring, excessive hardness, similar colour spot \\ \midrule
Fryum &
  \begin{tabular}[c]{@{}c@{}}irregular shape, unusual odor, uneven coloring, unusual texture, \\ small scratches, different colour spot, fryum stuck together, other\end{tabular} \\ \midrule
Macaroni1 &
  \begin{tabular}[c]{@{}c@{}}uneven shape , small scratches, small cracks, uneven coloring, \\ signs of insect infestation, uneven texture, Unusual consistency\end{tabular} \\ \midrule
Macaroni2 &
  \begin{tabular}[c]{@{}c@{}}irregular shape, small scratches, presence of foreign particles, \\ excessive moisture, signs of infestation, small cracks, unusual texture\end{tabular} \\ \midrule
Pcb1 &
  \begin{tabular}[c]{@{}c@{}}oxidation on the copper traces, separation of layers, presence of solder bridges, \\ excessive solder residue, discoloration, Uneven solder joints, bowing of the board, missing vias\end{tabular} \\ \midrule
Pcb2 &
  \begin{tabular}[c]{@{}c@{}}oxidation on the copper traces, separation of layers, presence of solder bridges, \\ excessive solder residue, discoloration, Uneven solder joints, bowing of the board, missing vias\end{tabular} \\ \midrule
Pcb3 &
  \begin{tabular}[c]{@{}c@{}}oxidation on the copper traces, separation of layers, presence of solder bridges, \\ excessive solder residue, discoloration, Uneven solder joints, bowing of the board, missing vias\end{tabular} \\ \midrule
Pcb4 &
  \begin{tabular}[c]{@{}c@{}}oxidation on the copper traces, separation of layers, presence of solder bridges, \\ excessive solder residue, discoloration, Uneven solder joints, bowing of the board, missing vias\end{tabular} \\ \midrule
Pipe\_fryum &
  \begin{tabular}[c]{@{}c@{}}uneven shape, presence of foreign objects, different colour spot, unusual odor, \\ empty interior, unusual texture, similar colour spot, stuck together\end{tabular} \\ \bottomrule
\end{tabular}
\caption{Fine-Grained anomaly description of every object within VisA dataset.}
\label{tab:detail-visa}
\end{table*}

In Figure~\ref{fig:detail_match}, we additionally display the similarity between each detailed anomaly description generated by LLM and the image features. We showcase the top 5 detailed anomaly descriptions with the highest similarity to the image, highlighting the most similar descriptions in red. By identifying the detailed anomaly description with the highest similarity, we can further discern the type of anomaly present in the sample.

\section{Additional Ablations}

In this section, we conducted further ablation studies on various detailed components of FiLo, including the backbone utilized, learning rate, employment of VV Attention, different treatments on QKV and VV Attention results, learning strategies for adaptive learning templates, number of learnable vectors, the structure and connectivity of Adapters, etc. Below are detailed analyses for each aspect.

\subsection{Different Backbones and Learning Rates}

Previous anomaly detection methods based on CLIP have typically utilized different CLIP backbones. WinCLIP\cite{jeong2023winclip} employs ViT-B-16@240px, while methods like APRIL-GAN~\cite{chen2023zero} and AnomalyCLIP~\cite{zhou2023anomalyclip} use ViT-L-14@336px. Existing methods have shown that using a backbone with higher image resolution is more beneficial for pixel-level anomaly localization. However, these methods with higher resolutions have not surpassed WinCLIP, which uses a resolution of 240x240, in terms of image-level AUC. We also implemented our FiLo method on these two commonly used backbones, and the results are shown in Table~\ref{tab:lr}.

In addition to the choice of backbone, the setting of learning rates also influences experimental results. Table~\ref{tab:lr} further illustrates the experimental results of FiLo under different learning rates ranging from 1e-3 to 1e-5. It can be observed that FiLo achieves the best anomaly detection and localization performance on both datasets when using a learning rate of 1e-3 for the learnable text vectors and a learning rate of 1e-4 for the MMCI module.

\begin{table*}[]
\centering

\begin{tabular}{@{}ccccccc@{}}
\toprule
\multirow{2}{*}{Backbone} & \multirow{2}{*}{learnable vec's lr} & \multirow{2}{*}{MMCI's lr} & \multicolumn{2}{c}{VisA} & \multicolumn{2}{c}{MVTec-AD} \\ \cmidrule(l){4-7} 
             &      &      & Image-AUC & Pixel-AUC & Image-AUC & Pixel-AUC \\ \midrule
ViT-B-16@240 & 1e-3 & 1e-4 & 78.1      & 93.5      & 77.9      & 88.2      \\
\textbf{ViT-L-14@336} & \textbf{1e-3} & \textbf{1e-4} & \textbf{83.9}      & \textbf{95.9}      & \textbf{91.2}      & \textbf{92.3}      \\
ViT-L-14@336 & 1e-3 & 1e-3 & 80.3      & 95.7      & 86.2      & 89.7      \\
ViT-L-14@336 & 1e-4 & 1e-4 & 82.4      & 95.7      & 88        & 91.2      \\
ViT-L-14@336 & 1e-4 & 1e-5 & 78.2      & 95.1      & 83.5      & 89        \\
ViT-L-14@336 & 1e-5 & 1e-5 & 80.4      & 95.2      & 85.8       &  90.7         \\ \bottomrule
\end{tabular}
\caption{Experimental results of FiLo on MVTec and VisA datasets under different backbones and learning rates.}
\label{tab:lr}
\end{table*}

\subsection{Adaptively Learned Text Templates}

CoOp~\cite{zhou2022learning} and CoCoOp~\cite{zhou2022conditional} are two distinct methods that utilize learnable vectors to replace manually crafted text prompts. These methods exhibit some differences in their approaches. Specifically, the learnable vectors in CoOp are agnostic to image content and are directly embedded into the text prompt, emphasizing the universality and uniformity of the text prompt. On the other hand, CoCoOp builds upon the learnable vectors embedded in the text prompt by incorporating a lightweight meta-net to append image features to the text prompt. This approach emphasizes generating tailored text prompts for each image, aiming to better match the image content.

Table~\ref{tab:coop} presents the experimental results of FiLo under the respective usage of CoOp and CoCoOp. Inspired by AnomalyCLIP~\cite{zhou2023anomalyclip}, we also explored the performance under the addition of class name information in the text content. The experimental results indicate that when using CoOp, omitting class name from the text yields better results, consistent with findings in AnomalyCLIP. This is because CoOp inherently emphasizes the generality and uniformity of the text prompt. Conversely, when employing CoCoOp for learning text templates, including class name information improves performance. This is attributed to the alignment of CoCoOp's approach, which incorporates image features into the text prompt via a meta-net, with the concept of FiLo, utilizing fine-grained anomaly description and position enhancement to obtain precise representations of each image's text content, aiming for a better match with image content.

The results in Table~\ref{tab:coop} further demonstrate that CoCoOp outperforms CoOp, highlighting the effectiveness of leveraging fine-grained anomaly descriptions to enhance anomaly detection.

\begin{table*}[]
\centering

\begin{tabular}{cccccc}
\hline
\multirow{2}{*}{learning method} & \multirow{2}{*}{with class name} & \multicolumn{2}{c}{VisA} & \multicolumn{2}{c}{MVTec-AD} \\ \cline{3-6} 
       &             & Image-AUC & Pixel-AUC & Image-AUC & Pixel-AUC \\ \hline
CoOp   &             & 81.7      & 95        & 90.8      & 89.5      \\
CoOp   & $\checkmark$ & 80.9      & 95.5      & 89.9      & 90.4      \\
CoCoOp &             & 82.3      & 95.4      & 91        & 90.5      \\
\textbf{CoCoOp} & $\checkmark$ & \textbf{83.9}      & \textbf{95.9}      & \textbf{91.2}      & \textbf{92.3}      \\ \hline
\end{tabular}
\caption{Comparison of different learning methods for learnable vectors and whether to use class name.}
\label{tab:coop}
\end{table*}

\begin{figure*}[]
  \centering
  \includegraphics[width=0.7\textwidth]{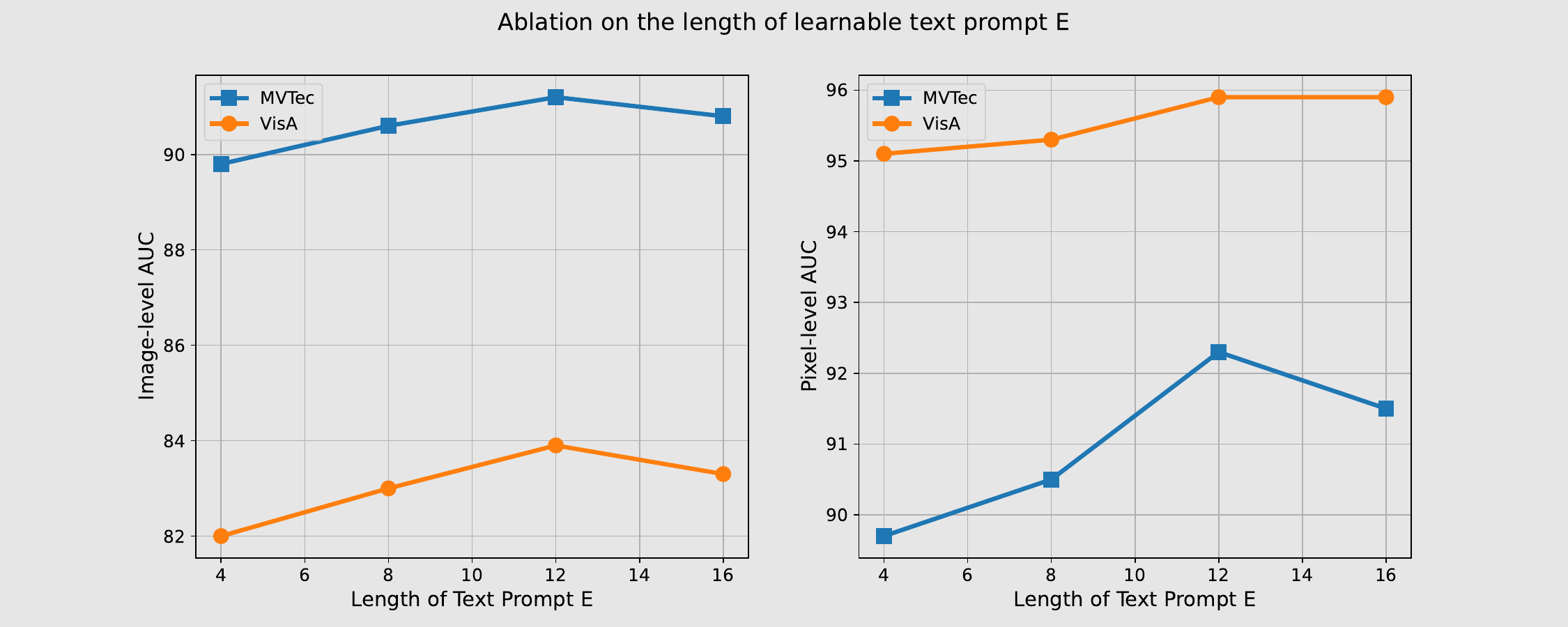} 
  \caption{Comparison of FiLo on MVTec and VisA datasets with different numbers of learnable vectors.}
  \label{fig:num_vec}
\end{figure*}

We also examined the impact of varying the number of learnable vectors in adaptively learned text templates. The findings are illustrated in Figure~\ref{fig:num_vec}. It can be observed that utilizing 12 learnable vectors yields the best performance in both anomaly detection and localization tasks.

\begin{table*}[]
\centering

\begin{tabular}{ccccccc}
\hline
\multirow{2}{*}{QKV results} & \multirow{2}{*}{VV results} & \multirow{2}{*}{VV's first layer} & \multicolumn{2}{c}{VisA} & \multicolumn{2}{c}{MVTec-AD} \\ \cline{4-7} 
       &        &   & Image-AUC & Pixel-AUC & Image-AUC & Pixel-AUC \\ \hline
Linear & MMCI   & 1 & 81.9      & 95.3      & 87.8      & 89.2      \\
\textbf{Linear} & \textbf{MMCI}   & \textbf{7} & \textbf{83.9}      & \textbf{95.9}      & \textbf{91.2}      & \textbf{92.3}      \\
Linear & Linear & 7 & 82.7      & 95.1      & 88.5      & 88.8      \\
MMCI   & MMCI   & 7 & 83.2      & 95.5      & 50        & 50        \\
MMCI   & Linear & 7 & 82.5      & 95.7      & 56.9      & 57.6      \\ \hline
\end{tabular}
\caption{Comparison of results of different processing methods for the output results of QKV and VV Attention.}
\label{tab:qkvvv}
\end{table*}

\begin{table*}[]
\centering
\begin{tabular}{@{}cccccc@{}}
\toprule
\multirow{2}{*}{Adapter's arch} & \multirow{2}{*}{residual} & \multicolumn{2}{c}{VisA} & \multicolumn{2}{c}{MVTec-AD} \\ \cmidrule(l){3-6} 
           &          & Image-AUC & Pixel-AUC & Image-AUC & Pixel-AUC \\ \midrule
\textbf{Bottleneck} &  & \textbf{83.9}      & \textbf{95.9}      & \textbf{91.2}      & \textbf{92.3}      \\
Bottleneck & $\checkmark$ & 83.6      & 95.8      & 89.9      & 91.4      \\
Linear     &  & 83.9      & 95.9      & 90.2      & 92.3      \\
Linear     & $\checkmark$ & 83.8      & 95.9      & 88.6      & 91.1      \\ \bottomrule
\end{tabular}

\caption{Comparison of different adapter structures and connection types.}
\label{tab:adapter}
\end{table*}

\subsection{Utilization of V-V Attention}

Pre-trained on large-scale datasets, CLIP exhibits excellent zero-shot performance on downstream image classification tasks. However, directly using the features extracted from the CLIP Image Encoder for each position in the feature map and measuring their similarity with textual features often results in significant noise activation outside of objects during fine-grained semantic segmentation or object detection tasks. CLIP Surgery~\cite{li2023clip} addresses this issue, identifying it as stemming from the QKV attention mechanism within CLIP, which leads to feature pooling from semantically disparate regions, consequently causing noise activation in erroneous areas. The proposed solution involves employing V-V self-attention to mitigate this problem.

Approaches such as AnoVL~\cite{deng2023anovl} and AnomalyCLIP~\cite{zhou2023anomalyclip} have also incorporated V-V attention into anomaly detection and localization tasks, resolving the issue of misalignment between patch-level features and textual features encountered in WinCLIP and APRIL-GAN, achieving remarkable zero-shot performance. However, V-V attention suffers from training difficulties, as slight mishandling may result in model outputs entirely comprised of zeros, causing the AUC to plummet to 50. To address this challenge, we simultaneously utilize the output results of both QKV attention and V-V attention, exploring the differential effects of applying distinct processing methods to the output results of QKV attention and V-V attention. The results, as shown in Table~\ref{tab:qkvvv}, indicate that employing a simple linear layer on the output results of QKV attention and inputting the output results of V-V attention into the MMCI module yields the best detection and localization performance for FiLo.

\begin{figure*}[]
  \centering
  \includegraphics[width=0.75\textwidth]{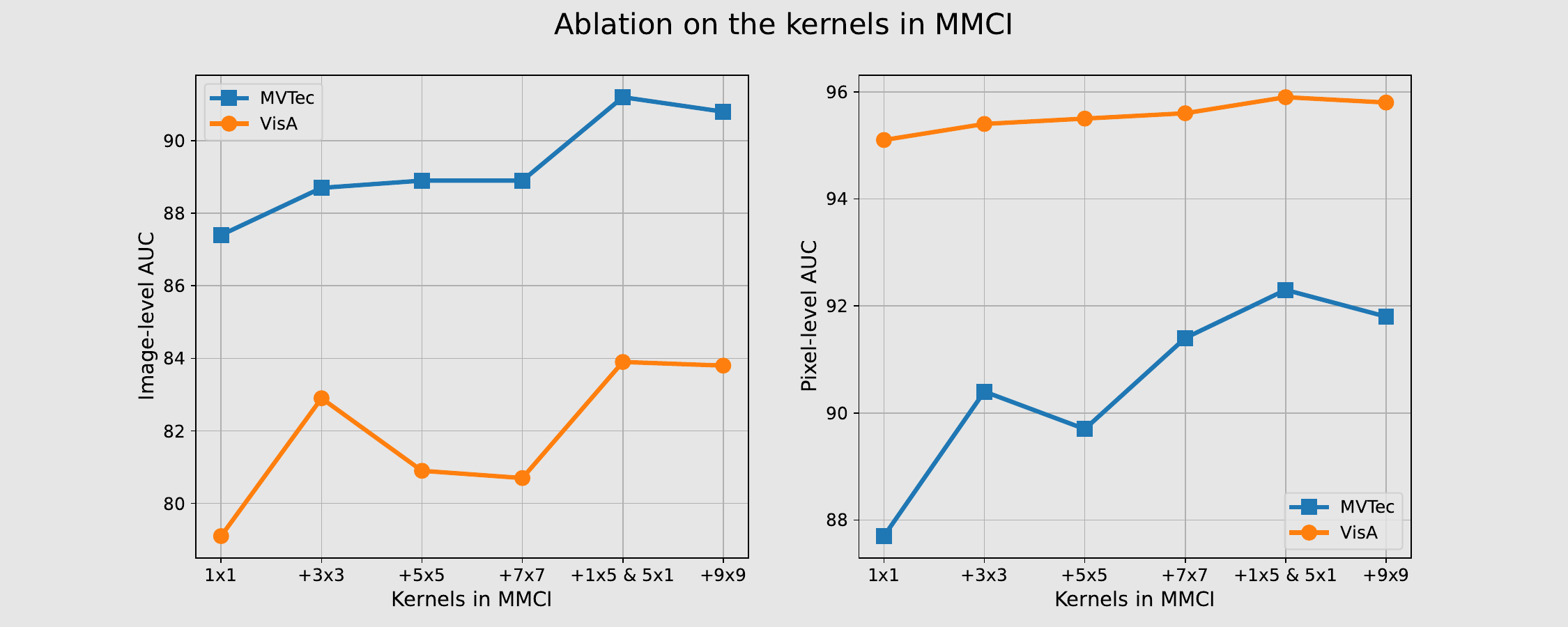} 
  \caption{Comparison of FiLo on MVTec and VisA datasets with different convolution kernels.}
  \label{fig:mmci_kernel}
\end{figure*}

\begin{figure*}[]
  \centering
  \includegraphics[width=0.75\textwidth]{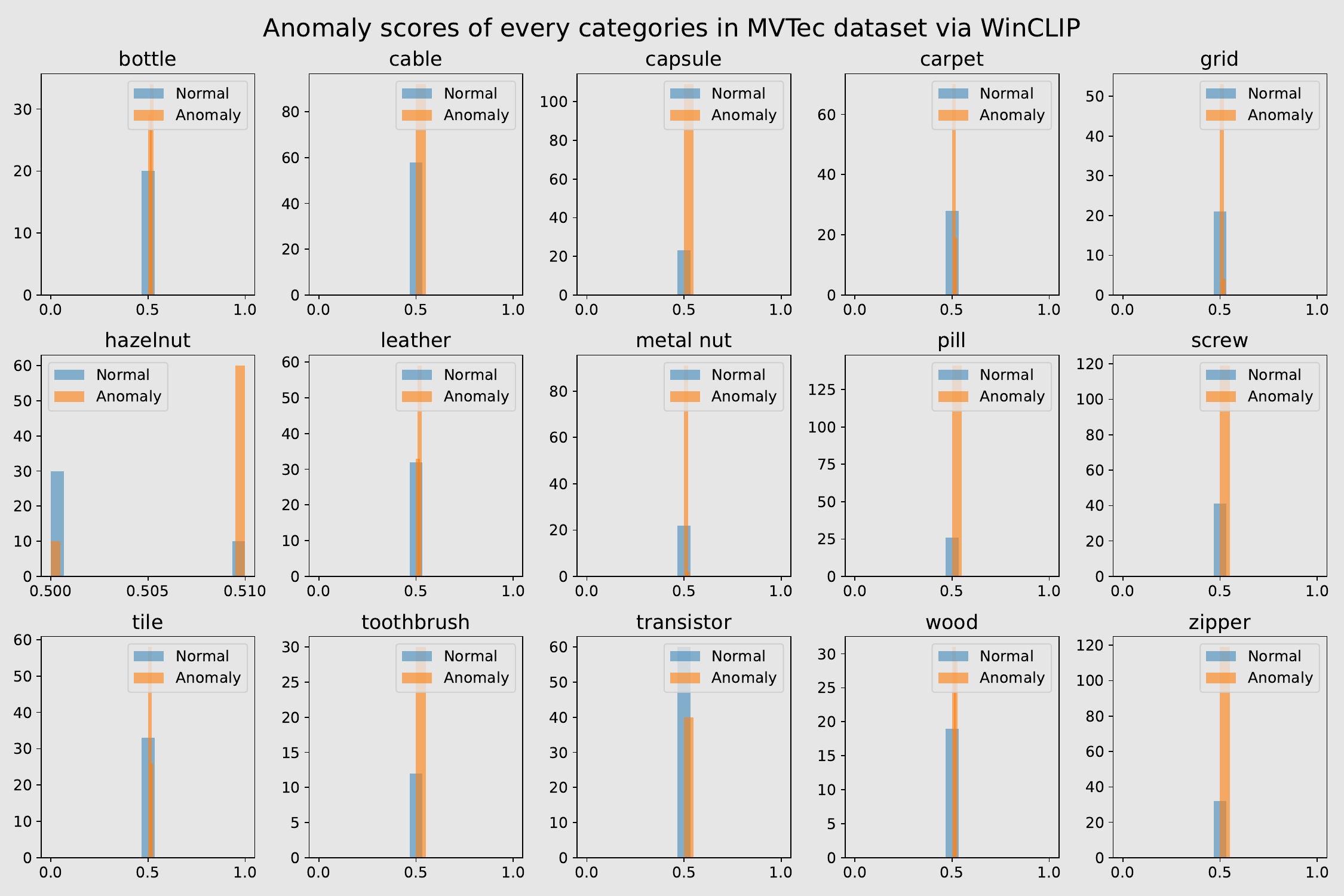} 
  \caption{Anomaly scores of WinCLIP on the MVTec dataset. Each sub-figure represents the visualization of one object.}
  \label{fig:scores_mvtec_winclip}
\end{figure*}

\begin{figure*}[]
  \centering
  \includegraphics[width=0.75\textwidth]{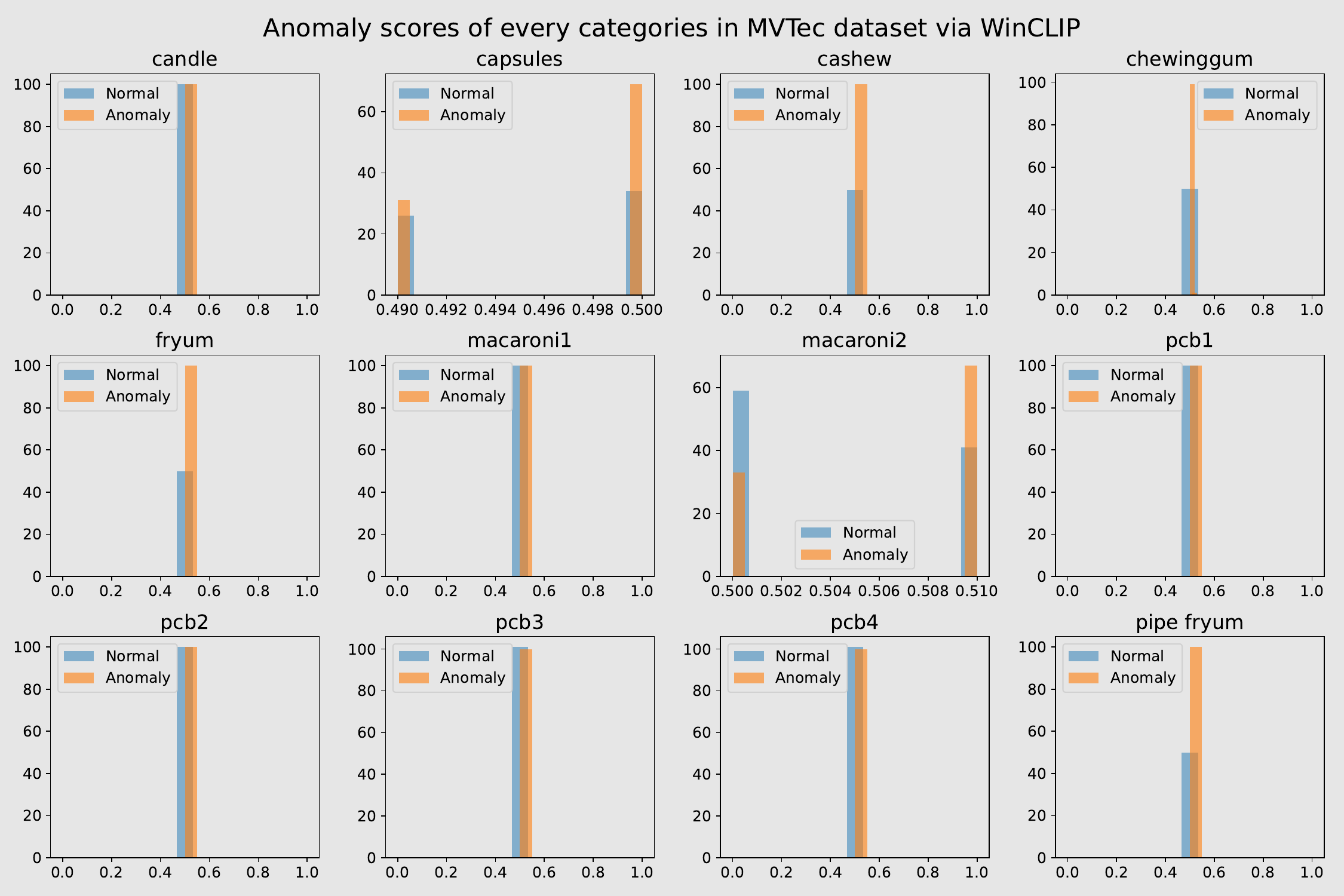} 
  \caption{Anomaly scores of WinCLIP on the VisA dataset. Each sub-figure represents the visualization of one object.}
  \label{fig:scores_visa_winclip}
\end{figure*}

\begin{figure*}[]
  \centering
  \includegraphics[width=0.75\textwidth]{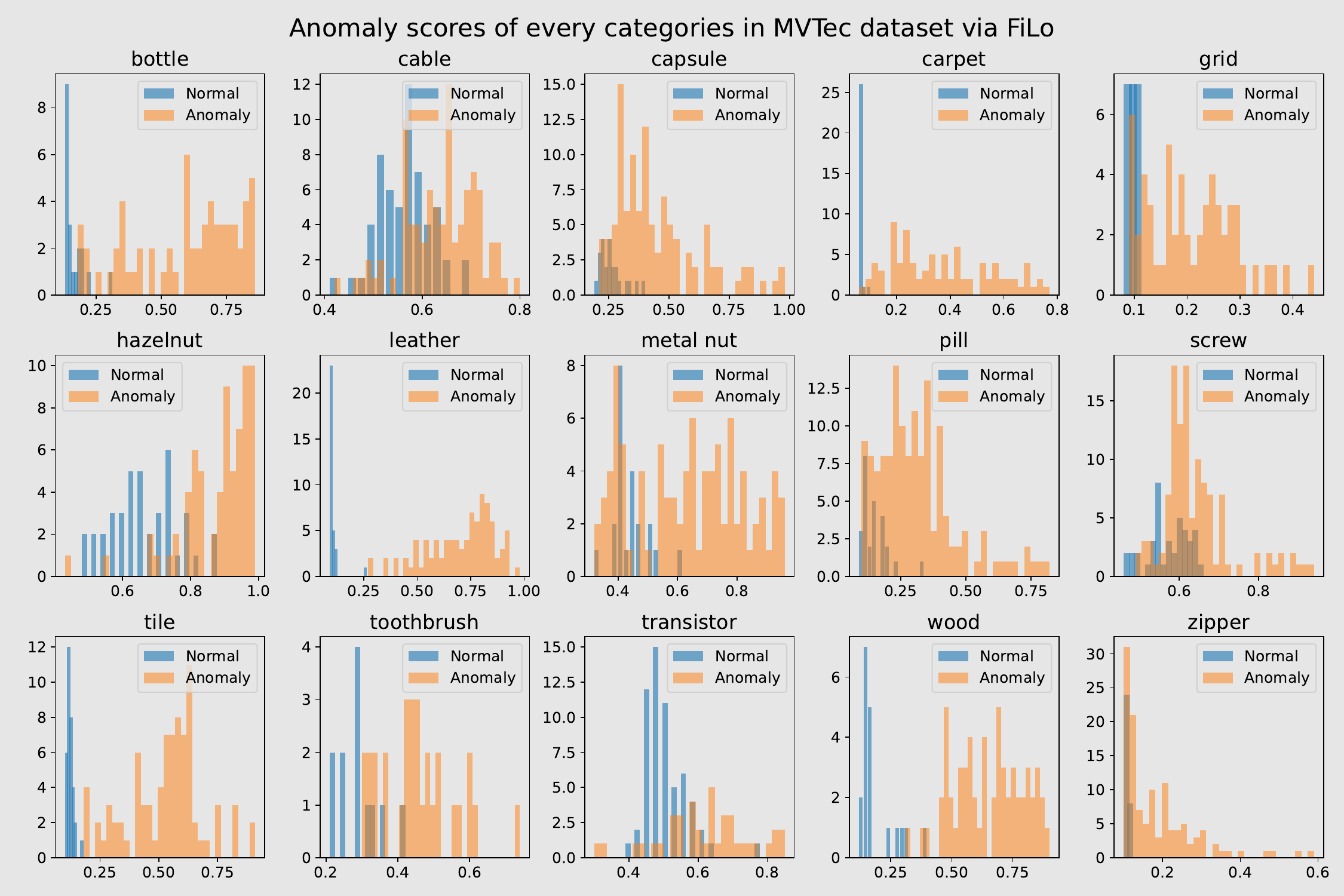} 
  \caption{Anomaly scores of FiLo on the MVTec dataset. Each sub-figure represents the visualization of one object.}
  \label{fig:scores_mvtec}
\end{figure*}

\begin{figure*}[]
  \centering
  \includegraphics[width=0.75\textwidth]{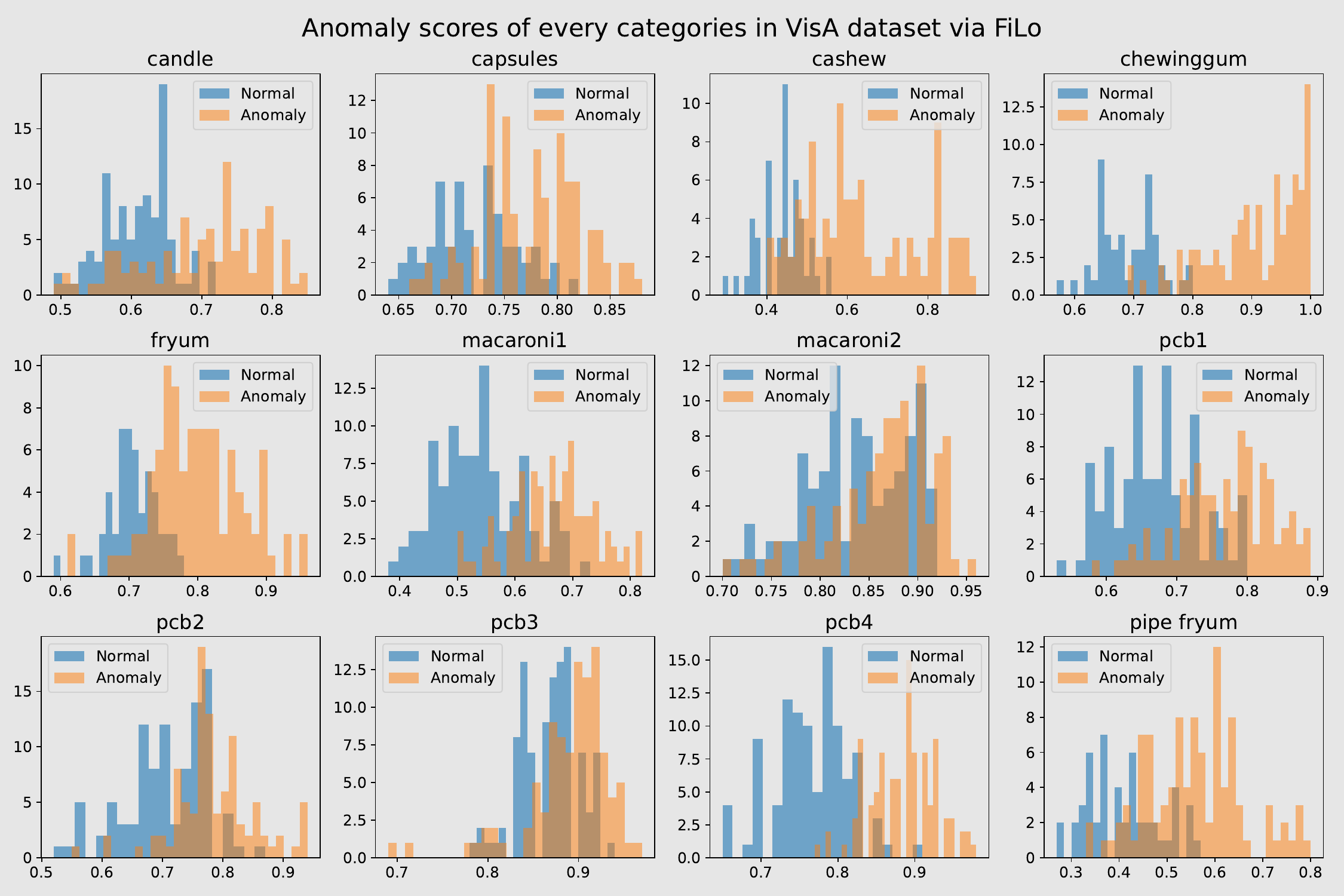} 
  \caption{Anomaly scores of FiLo on the VisA dataset. Each sub-figure represents the visualization of one object.}
  \label{fig:scores_visa}
\end{figure*}

\begin{figure*}[]
  \centering
  \includegraphics[width=0.8\textwidth]{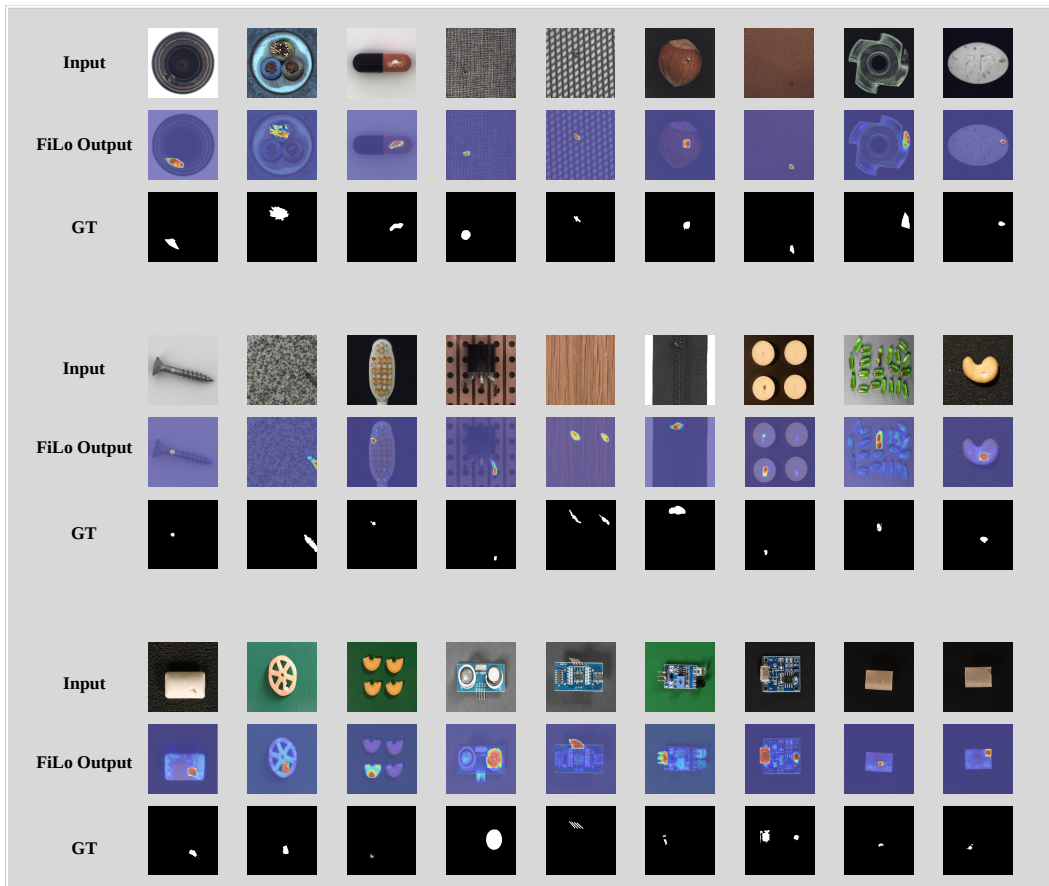} 
  \caption{More visualization results on MVTec and VisA datasets.}
  \label{fig:vis}
\end{figure*}

\subsection{Ablations of Adapter}

In this section, we compare the performance impact of the structure and connection methods of the adapter on FiLo. Regarding structure, we test the use of a simple linear layer and the bottleneck structure as shown in Sec 3 of the main paper. We also conduct experiments to assess the performance difference of the adapter when utilizing residual connection versus not utilizing it. Experimental results are shown in Table~\ref{tab:adapter}. It can be observed that when employing the bottleneck structure without residual connection, the adapter achieves the best performance.

\subsection{Convolution Kernel's Shape of MMCI}
\vspace{-1mm}

We extensively experiment on the impact of different kernel shapes used in MMCI. Starting with the sole use of 1x1 convolutional kernels and gradually incorporate shapes such as 3x3, 5x5, 7x7, 1x5, 5x1, and 9x9, we evaluate the various experimental results, as depicted in Figure~\ref{fig:mmci_kernel}. Based on the experimental findings, we ultimately select a combination of kernel shapes including 1x1, 3x3, 5x5, 7x7, 1x5, and 5x1. This combination harnesses the advantages of multi-scale and multi-shape kernels, enabling precise localization of anomalous regions of different sizes and shapes.

\section{Visualization}
\subsection{Anomaly Scores for Every Categories}

In this section, we present the statistical analysis of anomaly scores generated by WinCLIP~\cite{jeong2023winclip} and FiLo for each class object in the MVTec and VisA datasets. These visualizations aim to illustrate the effectiveness of FiLo's detailed anomaly descriptions and adaptively learned text templates compared to WinCLIP's manually crafted two-class text adjustment. As depicted in Figure~\ref{fig:scores_mvtec_winclip} and Figure~\ref{fig:scores_visa_winclip}, WinCLIP's scores for both normal and abnormal samples heavily overlap and are concentrated around 0.5, indicating its failure to effectively distinguish between normal and abnormal samples. In contrast, Figure~\ref{fig:scores_mvtec} and Figure~\ref{fig:scores_visa} illustrate FiLo's visualization results on these two datasets. It can be observed that the scores for normal samples significantly decrease while those for abnormal samples notably increase, resulting in a significant reduction in the overlapping area.

\subsection{Anomaly Maps}

Figure~\ref{fig:vis} further demonstrates the Anomaly Maps generated by FiLo on additional samples from the MVTec and VisA datasets. The three rows from top to bottom in the figure represent the test samples, FiLo's output, and the Ground Truth, respectively, demonstrating FiLo's robust anomaly localization capability.
% FiLo achieves results very close to the ground truth even under the 0-shot setting, showcasing its powerful anomaly localization capability.

% Figure~\ref{fig:vis} showcase the Anomaly Maps generated by FiLo on a broader set of samples from the MVTec and VisA datasets. The three lines from top to bottom in the figure are test samples, FiLo’s output results and Ground Truth, 

\section{Limitation and future work}

Compared to previous works like WinCLIP~\cite{jeong2023winclip}, FiLo has made advancements in anomaly detection, localization, and interpretability through the use of Fine-Grained Description and High-Quality Localization methods. However, despite these strides forward, certain limitations still persist, warranting further investigation and refinement. As illustrated in Figure~\ref{fig:scores_mvtec} and Figure~\ref{fig:scores_visa}, while the differentiation between normal and abnormal samples is more distinct compared to previous methods, significant overlap still exists in certain categories such as zipper and metal nut. In the future, we plan to further improve the differentiation between normal and abnormal sample scores through approaches such as metric learning.

%% file: main.bib
@String(CVPR= {IEEE Conf. Comput. Vis. Pattern Recog.})

@String(AAAI = {AAAI})

@String(CVPR  = {CVPR})

@inproceedings{roth2022towards,
  title={Towards total recall in industrial anomaly detection},
  author={Roth, Karsten and Pemula, Latha and Zepeda, Joaquin and Sch{\"o}lkopf, Bernhard and Brox, Thomas and Gehler, Peter},
  booktitle={Proceedings of the IEEE/CVF Conference on Computer Vision and Pattern Recognition},
  pages={14318--14328},
  year={2022}
}

@article{you2022unified,
  title={A unified model for multi-class anomaly detection},
  author={You, Zhiyuan and Cui, Lei and Shen, Yujun and Yang, Kai and Lu, Xin and Zheng, Yu and Le, Xinyi},
  journal={Advances in Neural Information Processing Systems},
  volume={35},
  pages={4571--4584},
  year={2022}
}

@inproceedings{radford2021learning,
  title={Learning transferable visual models from natural language supervision},
  author={Radford, Alec and Kim, Jong Wook and Hallacy, Chris and Ramesh, Aditya and Goh, Gabriel and Agarwal, Sandhini and Sastry, Girish and Askell, Amanda and Mishkin, Pamela and Clark, Jack and others},
  booktitle={International conference on machine learning},
  pages={8748--8763},
  year={2021},
  organization={PMLR}
}

@inproceedings{jeong2023winclip,
  title={Winclip: Zero-/few-shot anomaly classification and segmentation},
  author={Jeong, Jongheon and Zou, Yang and Kim, Taewan and Zhang, Dongqing and Ravichandran, Avinash and Dabeer, Onkar},
  booktitle={Proceedings of the IEEE/CVF Conference on Computer Vision and Pattern Recognition},
  pages={19606--19616},
  year={2023}
}

@article{chen2023zero,
  title={A zero-/few-shot anomaly classification and segmentation method for cvpr 2023 vand workshop challenge tracks 1\&2: 1st place on zero-shot ad and 4th place on few-shot ad},
  author={Chen, Xuhai and Han, Yue and Zhang, Jiangning},
  journal={arXiv preprint arXiv:2305.17382},
  year={2023}
}

@inproceedings{gu2024anomalygpt,
  title={Anomalygpt: Detecting industrial anomalies using large vision-language models},
  author={Gu, Zhaopeng and Zhu, Bingke and Zhu, Guibo and Chen, Yingying and Tang, Ming and Wang, Jinqiao},
  booktitle={Proceedings of the AAAI Conference on Artificial Intelligence},
  volume={38},
  number={3},
  pages={1932--1940},
  year={2024}
}

@article{deng2023anovl,
  title={Anovl: Adapting vision-language models for unified zero-shot anomaly localization},
  author={Deng, Hanqiu and Zhang, Zhaoxiang and Bao, Jinan and Li, Xingyu},
  journal={arXiv preprint arXiv:2308.15939},
  year={2023}
}

@article{liu2023grounding,
  title={Grounding dino: Marrying dino with grounded pre-training for open-set object detection},
  author={Liu, Shilong and Zeng, Zhaoyang and Ren, Tianhe and Li, Feng and Zhang, Hao and Yang, Jie and Li, Chunyuan and Yang, Jianwei and Su, Hang and Zhu, Jun and others},
  journal={arXiv preprint arXiv:2303.05499},
  year={2023}
}

@article{zhou2022learning,
  title={Learning to prompt for vision-language models},
  author={Zhou, Kaiyang and Yang, Jingkang and Loy, Chen Change and Liu, Ziwei},
  journal={International Journal of Computer Vision},
  volume={130},
  number={9},
  pages={2337--2348},
  year={2022},
  publisher={Springer}
}

@inproceedings{zhou2022conditional,
  title={Conditional prompt learning for vision-language models},
  author={Zhou, Kaiyang and Yang, Jingkang and Loy, Chen Change and Liu, Ziwei},
  booktitle={Proceedings of the IEEE/CVF conference on computer vision and pattern recognition},
  pages={16816--16825},
  year={2022}
}

@article{vaswani2017attention,
  title={Attention is all you need},
  author={Vaswani, Ashish and Shazeer, Noam and Parmar, Niki and Uszkoreit, Jakob and Jones, Llion and Gomez, Aidan N and Kaiser, {\L}ukasz and Polosukhin, Illia},
  journal={Advances in neural information processing systems},
  volume={30},
  year={2017}
}

@inproceedings{esmaeilpour2022zero,
  title={Zero-shot out-of-distribution detection based on the pre-trained model clip},
  author={Esmaeilpour, Sepideh and Liu, Bing and Robertson, Eric and Shu, Lei},
  booktitle={Proceedings of the AAAI conference on artificial intelligence},
  volume={36},
  number={6},
  pages={6568--6576},
  year={2022}
}

@article{liznerski2022exposing,
  title={Exposing outlier exposure: What can be learned from few, one, and zero outlier images},
  author={Liznerski, Philipp and Ruff, Lukas and Vandermeulen, Robert A and Franks, Billy Joe and M{\"u}ller, Klaus-Robert and Kloft, Marius},
  journal={arXiv preprint arXiv:2205.11474},
  year={2022}
}

@article{zhou2023anomalyclip,
  title={Anomalyclip: Object-agnostic prompt learning for zero-shot anomaly detection},
  author={Zhou, Qihang and Pang, Guansong and Tian, Yu and He, Shibo and Chen, Jiming},
  journal={arXiv preprint arXiv:2310.18961},
  year={2023}
}

@article{achiam2023gpt,
  title={Gpt-4 technical report},
  author={Achiam, Josh and Adler, Steven and Agarwal, Sandhini and Ahmad, Lama and Akkaya, Ilge and Aleman, Florencia Leoni and Almeida, Diogo and Altenschmidt, Janko and Altman, Sam and Anadkat, Shyamal and others},
  journal={arXiv preprint arXiv:2303.08774},
  year={2023}
}

@article{ouyang2022training,
  title={Training language models to follow instructions with human feedback},
  author={Ouyang, Long and Wu, Jeffrey and Jiang, Xu and Almeida, Diogo and Wainwright, Carroll and Mishkin, Pamela and Zhang, Chong and Agarwal, Sandhini and Slama, Katarina and Ray, Alex and others},
  journal={Advances in neural information processing systems},
  volume={35},
  pages={27730--27744},
  year={2022}
}

@inproceedings{szegedy2015going,
  title={Going deeper with convolutions},
  author={Szegedy, Christian and Liu, Wei and Jia, Yangqing and Sermanet, Pierre and Reed, Scott and Anguelov, Dragomir and Erhan, Dumitru and Vanhoucke, Vincent and Rabinovich, Andrew},
  booktitle={Proceedings of the IEEE conference on computer vision and pattern recognition},
  pages={1--9},
  year={2015}
}

@article{tan2019mixconv,
  title={Mixconv: Mixed depthwise convolutional kernels},
  author={Tan, Mingxing and Le, Quoc V},
  journal={arXiv preprint arXiv:1907.09595},
  year={2019}
}

@inproceedings{ding2021repvgg,
  title={Repvgg: Making vgg-style convnets great again},
  author={Ding, Xiaohan and Zhang, Xiangyu and Ma, Ningning and Han, Jungong and Ding, Guiguang and Sun, Jian},
  booktitle={Proceedings of the IEEE/CVF conference on computer vision and pattern recognition},
  pages={13733--13742},
  year={2021}
}

@inproceedings{ding2019acnet,
  title={Acnet: Strengthening the kernel skeletons for powerful cnn via asymmetric convolution blocks},
  author={Ding, Xiaohan and Guo, Yuchen and Ding, Guiguang and Han, Jungong},
  booktitle={Proceedings of the IEEE/CVF international conference on computer vision},
  pages={1911--1920},
  year={2019}
}

@inproceedings{deng2022anomaly,
  title={Anomaly detection via reverse distillation from one-class embedding},
  author={Deng, Hanqiu and Li, Xingyu},
  booktitle={Proceedings of the IEEE/CVF Conference on Computer Vision and Pattern Recognition},
  pages={9737--9746},
  year={2022}
}

@inproceedings{defard2021padim,
  title={Padim: a patch distribution modeling framework for anomaly detection and localization},
  author={Defard, Thomas and Setkov, Aleksandr and Loesch, Angelique and Audigier, Romaric},
  booktitle={International Conference on Pattern Recognition},
  pages={475--489},
  year={2021},
  organization={Springer}
}

@inproceedings{kirillov2023segment,
  title={Segment anything},
  author={Kirillov, Alexander and Mintun, Eric and Ravi, Nikhila and Mao, Hanzi and Rolland, Chloe and Gustafson, Laura and Xiao, Tete and Whitehead, Spencer and Berg, Alexander C and Lo, Wan-Yen and others},
  booktitle={Proceedings of the IEEE/CVF International Conference on Computer Vision},
  pages={4015--4026},
  year={2023}
}

@inproceedings{li2023blip,
  title={Blip-2: Bootstrapping language-image pre-training with frozen image encoders and large language models},
  author={Li, Junnan and Li, Dongxu and Savarese, Silvio and Hoi, Steven},
  booktitle={International conference on machine learning},
  pages={19730--19742},
  year={2023},
  organization={PMLR}
}

@inproceedings{bergmann2019mvtec,
  title={MVTec AD--A comprehensive real-world dataset for unsupervised anomaly detection},
  author={Bergmann, Paul and Fauser, Michael and Sattlegger, David and Steger, Carsten},
  booktitle={Proceedings of the IEEE/CVF conference on computer vision and pattern recognition},
  pages={9592--9600},
  year={2019}
}

@inproceedings{zou2022spot,
  title={Spot-the-difference self-supervised pre-training for anomaly detection and segmentation},
  author={Zou, Yang and Jeong, Jongheon and Pemula, Latha and Zhang, Dongqing and Dabeer, Onkar},
  booktitle={European Conference on Computer Vision},
  pages={392--408},
  year={2022},
  organization={Springer}
}

@article{cao2023segment,
  title={Segment any anomaly without training via hybrid prompt regularization},
  author={Cao, Yunkang and Xu, Xiaohao and Sun, Chen and Cheng, Yuqi and Du, Zongwei and Gao, Liang and Shen, Weiming},
  journal={arXiv preprint arXiv:2305.10724},
  year={2023}
}

@article{li2023clip,
  title={Clip surgery for better explainability with enhancement in open-vocabulary tasks},
  author={Li, Yi and Wang, Hualiang and Duan, Yiqun and Li, Xiaomeng},
  journal={arXiv preprint arXiv:2304.05653},
  year={2023}
}

@article{ren2024grounded,
  title={Grounded sam: Assembling open-world models for diverse visual tasks},
  author={Ren, Tianhe and Liu, Shilong and Zeng, Ailing and Lin, Jing and Li, Kunchang and Cao, He and Chen, Jiayu and Huang, Xinyu and Chen, Yukang and Yan, Feng and others},
  journal={arXiv preprint arXiv:2401.14159},
  year={2024}
}

@inproceedings{deng2009imagenet,
  title={Imagenet: A large-scale hierarchical image database},
  author={Deng, Jia and Dong, Wei and Socher, Richard and Li, Li-Jia and Li, Kai and Fei-Fei, Li},
  booktitle={2009 IEEE conference on computer vision and pattern recognition},
  pages={248--255},
  year={2009},
  organization={Ieee}
}

@inproceedings{lin2017focal,
  title={Focal loss for dense object detection},
  author={Lin, Tsung-Yi and Goyal, Priya and Girshick, Ross and He, Kaiming and Doll{\'a}r, Piotr},
  booktitle={Proceedings of the IEEE international conference on computer vision},
  pages={2980--2988},
  year={2017}
}

@inproceedings{milletari2016v,
  title={V-net: Fully convolutional neural networks for volumetric medical image segmentation},
  author={Milletari, Fausto and Navab, Nassir and Ahmadi, Seyed-Ahmad},
  booktitle={2016 fourth international conference on 3D vision (3DV)},
  pages={565--571},
  year={2016},
  organization={Ieee}
}

@article{loshchilov2017decoupled,
  title={Decoupled weight decay regularization},
  author={Loshchilov, Ilya and Hutter, Frank},
  journal={arXiv preprint arXiv:1711.05101},
  year={2017}
}

@inproceedings{glorot2011deep,
  title={Deep sparse rectifier neural networks},
  author={Glorot, Xavier and Bordes, Antoine and Bengio, Yoshua},
  booktitle={Proceedings of the fourteenth international conference on artificial intelligence and statistics},
  pages={315--323},
  year={2011},
  organization={JMLR Workshop and Conference Proceedings}
}

@article{elfwing2018sigmoid,
  title={Sigmoid-weighted linear units for neural network function approximation in reinforcement learning},
  author={Elfwing, Stefan and Uchibe, Eiji and Doya, Kenji},
  journal={Neural networks},
  volume={107},
  pages={3--11},
  year={2018},
  publisher={Elsevier}
}

@inproceedings{maniparambil2023enhancing,
  title={Enhancing clip with gpt-4: Harnessing visual descriptions as prompts},
  author={Maniparambil, Mayug and Vorster, Chris and Molloy, Derek and Murphy, Noel and McGuinness, Kevin and O'Connor, Noel E},
  booktitle={Proceedings of the IEEE/CVF International Conference on Computer Vision},
  pages={262--271},
  year={2023}
}

@article{menon2022visual,
  title={Visual classification via description from large language models},
  author={Menon, Sachit and Vondrick, Carl},
  journal={arXiv preprint arXiv:2210.07183},
  year={2022}
}

@article{feng2023leveraging,
  title={Leveraging multiple descriptive features for robust few-shot image learning},
  author={Feng, Zhili and Bair, Anna and Kolter, J Zico},
  journal={arXiv preprint arXiv:2307.04317},
  year={2023}
}
